\documentclass{article} 
\usepackage{iclr2027_conference,times}

\usepackage{amsmath,amsfonts,bm}

\def\eqref#1{equation~\ref{#1}}

\def\ceil#1{\lceil #1 \rceil}

\def\1{\bm{1}}

\def\eps{{\epsilon}}

\def\vf{{\bm{f}}}

\def\vk{{\bm{k}}}

\def\vm{{\bm{m}}}

\def\vo{{\bm{o}}}

\def\vq{{\bm{q}}}

\def\vv{{\bm{v}}}

\def\vx{{\bm{x}}}

\def\vz{{\bm{z}}}

\def\evalpha{{\alpha}}

\def\evlambda{{\lambda}}
\def\evomega{{\omega}}

\def\evtheta{{\theta}}

\def\evk{{k}}

\def\evq{{q}}

\def\mA{{\bm{A}}}

\def\mD{{\bm{D}}}

\def\mI{{\bm{I}}}

\def\mL{{\bm{L}}}

\def\mQ{{\bm{Q}}}
\def\mR{{\bm{R}}}

\def\mV{{\bm{V}}}
\def\mW{{\bm{W}}}

\def\mZ{{\bm{Z}}}

\def\mPhi{{\bm{\Phi}}}
\def\mLambda{{\bm{\Lambda}}}

\DeclareMathAlphabet{\mathsfit}{\encodingdefault}{\sfdefault}{m}{sl}
\SetMathAlphabet{\mathsfit}{bold}{\encodingdefault}{\sfdefault}{bx}{n}
\newcommand{\tens}[1]{\bm{\mathsfit{#1}}}

\def\tI{{\tens{I}}}

\def\gA{{\mathcal{A}}}

\def\gE{{\mathcal{E}}}
\def\gF{{\mathcal{F}}}
\def\gG{{\mathcal{G}}}
\def\gH{{\mathcal{H}}}
\def\gI{{\mathcal{I}}}

\def\gM{{\mathcal{M}}}
\def\gN{{\mathcal{N}}}

\def\gT{{\mathcal{T}}}

\def\gV{{\mathcal{V}}}

\newcommand{\Ls}{\mathcal{L}}
\newcommand{\R}{\mathbb{R}}

\newcommand{\reg}{\lambda}

\newcommand{\softmax}{\mathrm{softmax}}

\usepackage{hyperref}
\usepackage{url}
\usepackage{enumitem}

\usepackage{booktabs}
\usepackage{multirow}
\usepackage{graphicx}
\usepackage[table]{xcolor}

\title{TRoPE: Topology-aware Relative Positional Encoding for 4D Generation}

\author{%
Chengzhuo Gui \quad Zhenglin Zhou \quad Fan Ma \quad
Zongxin Yang \quad Yi Yang \\
Zhejiang University \\
{\hypersetup{pdfborder={0 0 0}}%
\href{https://akui1321.github.io/TRoPE}{%
\textcolor{projectblue}{\texttt{https://akui1321.github.io/TRoPE}}}}
}

\iclrfinalcopy 
\begin{document}

\maketitle

\begin{figure}[h]
    \centering
    \includegraphics[width=1\linewidth]{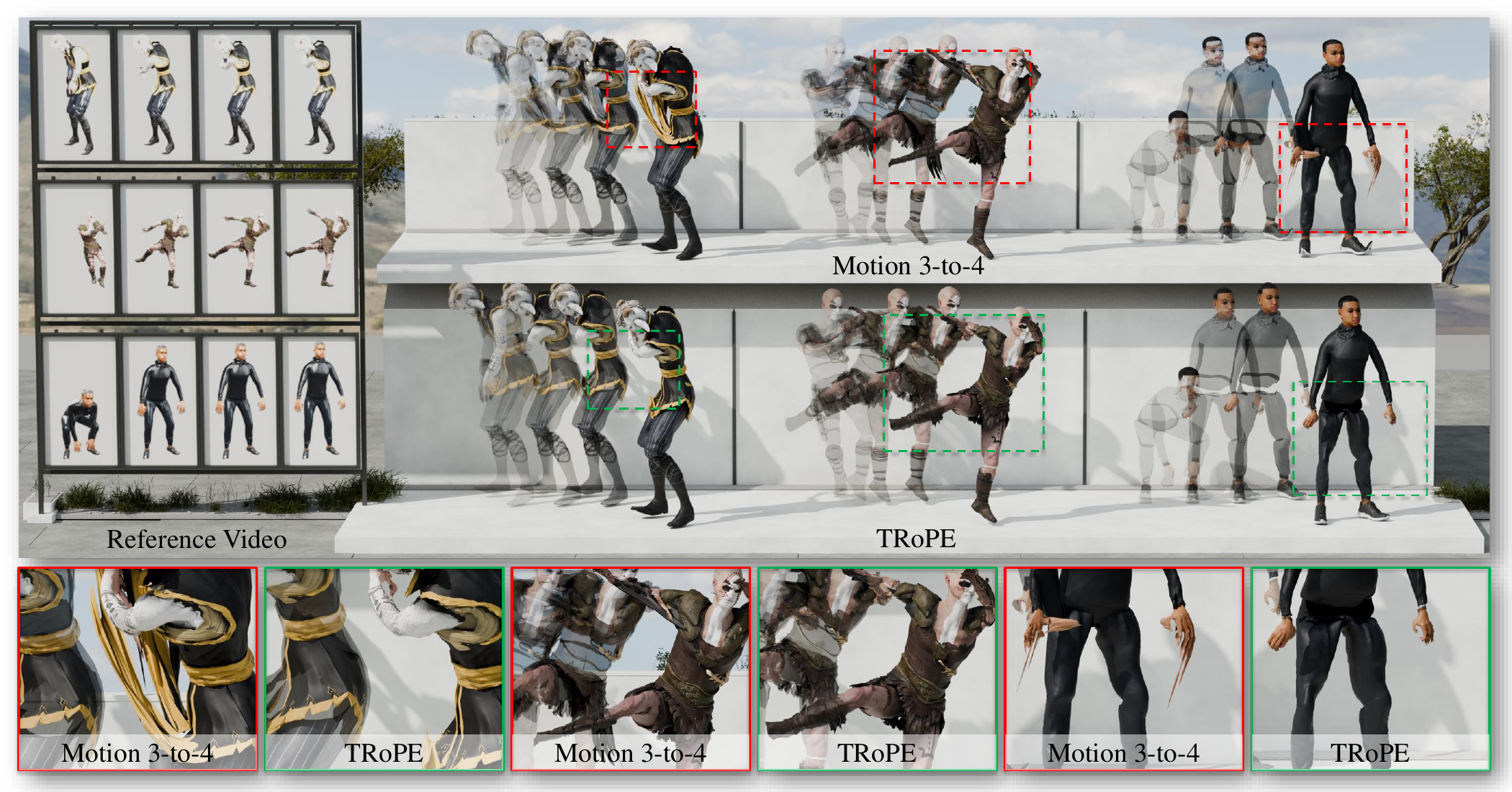}
    \caption{\textbf{TRoPE reduces sticking artifacts in 4D generation.}
    Given a reference mesh and a monocular video (left \added{panel}),
    \rev{the upper and lower platforms present the Motion 3-to-4 baseline and TRoPE, respectively. The bottom insets show zoomed-in comparisons of corresponding regions.}
    Compared with the baseline, our method
    reduces sticking artifacts between nearby
    body parts, producing more faithful motion.
    }
\label{fig:teaser}
\end{figure}

\begin{abstract}
Video-to-4D mesh generation has made substantial progress in recovering dynamic 3D geometry and motion from monocular videos.
However, commonly used feed-forward approaches remain prone to \textbf{sticking artifacts}, where geodesically distant vertices move together incorrectly when they become spatially close.
We attribute this failure to \textbf{cross-part motion leakage}: spatially nearby yet structurally distant vertices may retrieve similar motion information when motion decoding fails to \textit{distinguish spatial proximity from mesh connectivity}.
To address this issue, we introduce \textbf{TRoPE}, a topology-aware relative positional encoding that \rev{makes mesh connectivity explicit in the interaction between vertex queries and mesh-supported motion sources}.
TRoPE derives spectral features from the graph Laplacian and constructs pairwise topological relations from distances between multi-scale diffusion features, \rev{using these relations to modulate vertex--anchor attention through rotary phases}.
We further introduce a vertex-wise tail-risk objective that emphasizes persistently high-error vertices to explicitly suppress sticking artifacts.
To better characterize and quantify these failures, we introduce \textbf{MeshSticking-4D}, a dedicated benchmark for this failure mode.
\rev{Experiments on MeshSticking-4D show that TRoPE suppresses sticking artifacts while improving overall motion accuracy.}
Compared with the baseline, it reduces the worst-vertex error by 39.4\% and lowers the separation failure rate from 49.35\% to 34.90\%.
\rev{Additional evaluations on Motion-80 and Motion324-Subset show lower tail errors on diverse objects, while transfer to Mesh4D demonstrates applicability across backbones.}
\end{abstract}
\section{Introduction}

Video-to-4D mesh generation recovers dynamic 3D geometry and motion from monocular videos. Traditional pipelines bind vertices to a skeleton before driving the mesh; automatic-rigging methods such as Puppeteer~\citep{puppeteer} reduce this setup. Feed-forward methods, including Motion 3-to-4, R-DMesh, ActionMesh, and Mesh4D~\citep{motion324,rdmesh,actionmesh,mesh4d}, directly infer vertex trajectories or deformations, avoiding explicit rigging and enabling efficient animation of diverse meshes. This paradigm benefits applications such as game development, simulation, robotics, and world modeling.

Despite this progress, feed-forward models exhibit \textbf{sticking artifacts}: as shown in Figure~\ref{fig:teaser}, geodesically distant vertices may move together incorrectly when they become spatially close. Examples include hands sticking to the torso and garments remaining attached to nearby limbs. Such failures become more pronounced in long sequences and complex motions, limiting the practical use of these methods.

This behavior reflects \textbf{cross-part motion leakage} in the decoder. Feed-forward models compress video into motion tokens and let each dense vertex query them through pairwise attention. These vertices are typically parameterized by their spatial or geometric features. As a result, vertices from geodesically distant but spatially nearby regions may attend to similar motion latents. This causes motion to leak across unrelated parts and ultimately produces \textbf{sticking artifacts}.~\citep{motion324,animateanymesh++}.
\added{Topology-aware aggregation enriches individual vertex descriptors, while the structural relation between a query and its motion source remains implicit during motion retrieval.}
Because Euclidean proximity changes with articulation while reference-mesh connectivity remains fixed, motion decoding needs a persistent, connectivity-aware relative signal.
\added{Our central design is therefore to give motion sources mesh locations and make their structural relations to vertex queries explicit in attention.}

To this end, we introduce \textbf{TRoPE}, a topology-aware relative positional encoding that incorporates mesh connectivity into motion decoding.
\added{We first associate motion features with anchors sampled on the reference mesh, giving the motion sources explicit surface locations.}
\rev{We then} derive spectral features from the graph Laplacian to characterize structural relationships on the mesh, enabling motion retrieval to \textit{distinguish mesh connectivity from spatial proximity}. However, directly using spectral features across meshes is challenging due to the sign and eigenspace ambiguities of Laplacian eigenvectors~\citep{signnet_basisnet}.
\rev{We therefore} construct pairwise topological relations using distances between multi-scale diffusion features \rev{and use these relations to modulate vertex--anchor attention through rotary phases}.
Furthermore, we introduce a vertex-wise tail-risk measure that emphasizes vertices with persistently high errors and formulate it as a training objective to explicitly suppress sticking artifacts.

To evaluate our method, we introduce \textbf{MeshSticking-4D}, a benchmark centered on sticking artifacts.
\rev{Experiments on MeshSticking-4D show that TRoPE reduces sticking artifacts while improving overall motion accuracy, reducing the worst-vertex error by 39.4\% and lowering the separation failure rate from 49.35\% to 34.90\%.}
\rev{Additional evaluations on Motion-80~\citep{motion324} and Motion324-Subset show lower tail errors on diverse objects, while transfer experiments on Mesh4D demonstrate applicability across backbones.}

Our contributions are summarized as follows:

\begin{enumerate}[
    label=\textbullet,
    leftmargin=*,
    labelindent=0pt,
    align=left,
    itemsep=1.5pt,
    topsep=0pt,
    parsep=0pt
]
    \item We identify cross-part motion leakage as a key cause of sticking artifacts in feed-forward Video-to-4D generation, and highlight \rev{the need for explicit structural relations between vertex queries and motion sources}.

    \item We introduce TRoPE, a topology-aware relative positional encoding that \rev{makes mesh connectivity explicit in vertex--motion-source interactions}. TRoPE constructs \rev{relative topology relations from pairwise distances between multi-scale diffusion features and uses them to modulate vertex--anchor attention through rotary phases}.

    \item We introduce MeshSticking-4D, a benchmark specifically designed to evaluate sticking artifacts.
    \rev{Experiments demonstrate improved motion accuracy and sticking robustness on MeshSticking-4D, lower tail errors on Motion-80 and Motion324-Subset, and transferability to Mesh4D.}
\end{enumerate}
\section{Related Work}

\textbf{Absolute and Relative Positional Encodings.}
Transformer positional information has progressed from absolute token embeddings to pairwise relative representations~\citep{relandabs}.
Relative positions can be injected into attention via additive biases or SO(2)-based rotation~\citep{vaswani,relative,t5,roformer,alibi}.
In geometric vision, camera-conditioning methods such as CaPE, GTA, Repast, and PRoPE replace global camera coordinates with relative relationships between interacting views~\citep{CaPE,repast,gta,prope}.
Across multiple domains, positional encodings are shifting from absolute to relative formulations. In this work, we compare absolute and relative encodings for mesh representations to incorporate mesh connectivity into motion generation.

\textbf{Mesh Representations.}
Mesh connectivity provides structural relationships that are stable under deformation.
Classical spectral approaches to shape analysis, including the heat kernel signature, wave kernel signature, and functional maps, capture intrinsic relationships between surface locations~\citep{hks,hksbase,wks,functional,functionalmaps}.
Modern mesh and graph networks use adjacency, surface diffusion, graph shortest-path distances, or spectral features to model local and global structure, as demonstrated by DiffusionNet, MeshGraphNets, Mesh Graphormer, Graphormer, GraphiT, SAN, and GraphGPS~\citep{diffusionnet,meshgraphnets,meshgraphormer,graphormer,graphit,san,graphgps}.
SignNet and BasisNet further address sign and eigenspace ambiguities in spectral representations~\citep{signnet_basisnet}.
\added{MeshLoom~\citep{meshloom} uses topology-aware graph convolutions for non-rigid mesh-sequence registration and augments decoder keys and values with 3D Fourier positional features.}
These studies provide powerful tools for representing mesh structure.
In this work, we study a critical design decision: how to incorporate mesh connectivity into motion generation.

\textbf{Video-to-4D Generation.}
Video-to-4D generation has evolved from optimizing implicit or Gaussian representations with image and video priors to feed-forward models that explicitly reconstruct dynamic geometry.
Methods such as 4D-fy, Consistent4D, SV4D, STAG4D, and CAT4D exploit multi-view or video diffusion priors to synthesize temporally varying scenes, while L4GM and GVFD model dynamic objects with Gaussian or latent representations~\citep{fourdfy,consistent4d,sv4d,stag4d,cat4d,l4gm,gvfd}.
A parallel line focuses on explicit mesh representations. V2M4 and ShapeGen4D reconstruct meshes frame by frame and align them over time, whereas Motion 3-to-4, R-DMesh, Mesh4D, and ActionMesh learn explicit vertex trajectories or mesh deformations from video~\citep{v2m4,shapegen4d,motion324,rdmesh,mesh4d,actionmesh}.
\added{In mesh animation, AnimateAnyMesh and AnimateAnyMesh++~\citep{animateanymesh,animateanymesh++} enrich vertex features through topology-aware aggregation, a design also used by R-DMesh~\citep{rdmesh}.}
\rev{TRoPE complements topology-aware feature aggregation with an explicit connectivity-derived query--motion-source relation that modulates query--key compatibility through rotary phases during vertex--anchor motion retrieval.}

\section{Method}

We introduce \textbf{TRoPE}, a topology-aware relative positional encoding for 4D generation.
The host generator encodes the video into compact motion tokens, while TRoPE changes how mesh vertices retrieve those tokens.
A vertex-wise tail-risk objective emphasizes the persistent high-error vertices that form sticking artifacts.

\subsection{Video-to-4D Motion Decoding}
\label{sec:preliminaries}

Given a monocular video $\tI_{1:T}$ and a reference mesh
$\gM^{\mathrm{ref}}=(\mV^{\mathrm{ref}},\gF)$, video-to-4D generation
predicts a dynamic mesh sequence
\begin{equation}
\widehat{\gM}_{1:T}
=\bigl\{(\widehat{\mV}_t,\gF)\bigr\}_{t=1}^{T},
\qquad \widehat{\mV}_t\in\R^{N\times 3}.
\label{eq:mesh_animation_task}
\end{equation}
The vertex indexing and face connectivity of the reference mesh are fixed
over time, whereas Euclidean distances can change substantially during
articulation.

The host generator encodes the video and reference geometry into a compact set of
per-frame motion tokens $\mZ_t=\{\vz_{t,j}\}_{j=1}^{S}$, with $S\ll N$.
Each mesh vertex forms a query and retrieves motion through
content-based attention. For head $h$,
\begin{equation}
\ell_{itj}^{h}
=\frac{(\vq_i^{h})^{\top}\vk_{t,j}^{h}}{\sqrt{d_h}},
\qquad
p_{itj}^{h}=\softmax_{j}(\ell_{itj}^{h}),
\qquad
\vo_{i,t}^{h}=\sum_{j=1}^{S}p_{itj}^{h}\vv_{t,j}^{h}.
\label{eq:host_motion_attention}
\end{equation}
Thus, each vertex chooses motion information from the token set. When two
geodesically distant parts become close in the Euclidean space, their query
features can become similar and retrieve similar tokens. This
\emph{cross-part motion leakage} couples trajectories that should remain
separate and appears as \emph{sticking artifacts}. An absolute coordinate or
spectral code attached to each query does not specify the
relationship between that query and a motion token; raw spectral channels also depend on the arbitrary orientation of
the returned eigenbasis~\citep{signnet_basisnet}. We therefore give the motion source a mesh support
and encode the query--source relation directly.

\subsection{Mesh-supported Topological Relations}
\label{sec:mesh_relative_position}

\begin{figure*}
    \centering
    \includegraphics[width=1.0\linewidth]{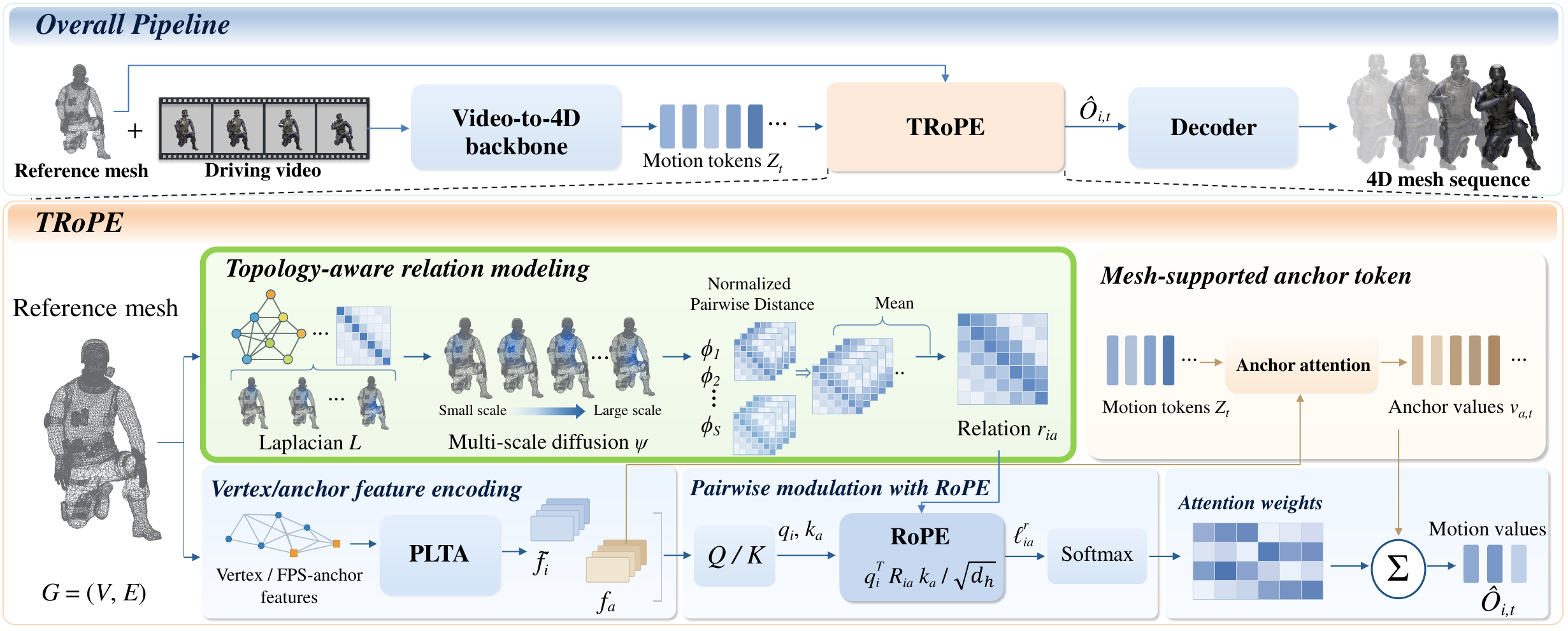}
    \vspace{-2.0em}
    \caption{
    \textbf{Overview of TRoPE for Topology-Aware Video-to-4D Mesh Generation.}
    Given a driving video and a reference mesh, a video-to-4D backbone extracts
    motion tokens, which TRoPE associates with mesh-supported anchors before a
    decoder generates the 4D mesh sequence.
    The lower panel details TRoPE:
    PLTA encodes local vertex and anchor features, while the reference
    connectivity graph is transformed through the graph Laplacian and
    multi-scale diffusion into pairwise topology-relative relations. These
    relations modulate query--key compatibility through rotary attention,
    producing motion features that preserve mesh structure and suppress
    cross-part motion leakage and sticking artifacts.
    }
    \label{fig:framework}
\end{figure*}

To distinguish mesh connectivity from spatial proximity during motion
decoding, we construct relative positions on the reference graph
$\gG=(\gV,\gE)$ induced by the mesh faces.

\textbf{Mesh-supported motion sources.}
To give motion features explicit mesh locations, we sample surface anchors
$\gA$ by farthest-point sampling. To incorporate local connectivity into
their interaction with vertex queries, we adapt the topology-aware
aggregation of AnimateAnyMesh++~\citep{animateanymesh++} into a local sparse
attention block, denoted PLTA:
\begin{equation}
\widetilde{\vf}_i
=\operatorname{PLTA}\bigl(\vf_i,\{\vf_j:j\in\gN_{\gG}(i)\}\bigr).
\label{eq:local_topology_features}
\end{equation}
Here, $\vf_i$ is the initial feature and $\gN_{\gG}(i)$ its mesh
neighborhood. The enriched anchor features query the host motion tokens:
\begin{equation}
\vm_{a,t}=\operatorname{Readout}(\widetilde{\vf}_a,\mZ_t),
\qquad a\in\gA.
\label{eq:anchor_motion_features}
\end{equation}
\added{This first stage uses content-based cross-attention: queries are the enriched anchor features, while keys and values are the per-frame host motion tokens.}
This associates each retrieved motion feature with a reference-surface
location, allowing vertex--anchor connectivity to inform subsequent attention.
\added{The host tokens themselves have no assigned reference-surface locations; anchors provide the mesh support needed to define the relation used in the subsequent vertex--anchor motion retrieval.}

\textbf{Multi-scale connectivity.}
To describe connectivity beyond local neighborhoods, we derive diffusion
features from the graph Laplacian~\citep{diffusionnet}. We process each connected component
independently to capture its internal structure. For component $c$,
\begin{equation}
\mL_c=\mI-\mD_c^{-1/2}\mA_c\mD_c^{-1/2}
=\mPhi_c\mLambda_c\mPhi_c^{\top}.
\label{eq:component_laplacian}
\end{equation}
where $\mA_c$ and $\mD_c$ are its adjacency and degree matrices.
We retain non-trivial modes indexed by $\gI_c$.
Let $\bar{\evlambda}_{c,m}$ denote eigenvalues normalized by the median
positive retained value, and $\widetilde{\bm{\phi}}_{c,m}$ the
component-standardized, clipped eigenvectors. To capture different
structural ranges, we construct features at diffusion scale $\tau$:
\begin{equation}
\begin{aligned}
\bm{\psi}_{c,\tau}(i)
&=\bigl(\sqrt{w_{\tau,c,m}}\,
\widetilde{\bm{\phi}}_{c,m}(i)\bigr)_{m\in\gI_c},\\[-1pt]
w_{\tau,c,m}
&=\frac{\exp\!\left(-\tau\bar{\evlambda}_{c,m}\right)}
{\left\|\bigl(\exp\!(-\tau\bar{\evlambda}_{c,r})\bigr)_{r\in\gI_c}\right\|_2}.
\end{aligned}
\label{eq:diffusion_factors}
\end{equation}
To express the relationship between a query and a motion source, we compare
their diffusion features through normalized distances across scales $\gT$.
For valid endpoints in the same component,
\begin{equation}
r_{ia}=\frac{1}{|\gT|}\sum_{\tau\in\gT}
\frac{\left\|\bm{\psi}_{c,\tau}(i)-\bm{\psi}_{c,\tau}(a)\right\|_2}
{\max\!\left\{
\left\|\bm{\psi}_{c,\tau}(i)\right\|_2+
\left\|\bm{\psi}_{c,\tau}(a)\right\|_2,\eps\right\}}.
\label{eq:mesh_relative_relation}
\end{equation}
\rev{The relation compares the structural neighborhoods of a query and its motion-source anchor across diffusion scales, rather than treating their spectral coordinates as independent positional inputs.}
\rev{Surface samples inherit spectral features by averaging the features at the three vertices of their containing triangle.}
\added{Components with fewer than 16 vertices are treated as invalid.} We assign $r_{ia}=0$ to cross-component or
invalid pairs, preserving content-based attention for those pairs.
To keep the structural information independent of articulation, we compute
the relation from reference connectivity and reuse it across frames.

\subsection{Topology-aware Relative RoPE}
\label{sec:topology_relative_rotary}

To make motion retrieval depend on both content and connectivity, we use
$r_{ia}$ to modulate query--key compatibility through a rotary phase~\citep{roformer}.
Rotation changes feature alignment while preserving feature norms.
For fixed rotary
frequencies $\{\evomega_f\}_{f=1}^{F_r}$ with $2F_r\leq d_h$,
\begin{equation}
\begin{gathered}
\evtheta_{ia,f}=2\pi\evomega_f r_{ia},
\qquad
\mR(\evtheta)=
\begin{bmatrix}\cos\evtheta&\sin\evtheta\\-\sin\evtheta&\cos\evtheta\end{bmatrix},\\
\mR_{ia}=\operatorname{diag}\bigl(
\mR(\evtheta_{ia,1}),\ldots,\mR(\evtheta_{ia,F_r}),
\mI_{d_h-2F_r}\bigr).
\end{gathered}
\label{eq:trope_rotation}
\end{equation}
\rev{In the second attention stage, queries} $\vq_i^h$ and keys $\vk_a^h$ are projected from the enriched
reference features $\widetilde{\vf}_i$ and $\widetilde{\vf}_a$.
We compute their compatibility as
\begin{equation}
\widetilde{\ell}_{ia}^{h}
=\frac{(\vq_i^{h})^{\top}\mR_{ia}\vk_a^{h}}{\sqrt{d_h}}.
\label{eq:trope_score}
\end{equation}
We use this form to let the same structural relation affect different
feature pairs differently. For one two-dimensional channel pair,
\begin{equation}
\vq^{\top}\mR(\evtheta)\vk
=(\evq_1\evk_1+\evq_2\evk_2)\cos\evtheta
+(\evq_1\evk_2-\evq_2\evk_1)\sin\evtheta.
\label{eq:trope_expansion}
\end{equation}
Connectivity sets the cosine and sine coefficients, while the learned
features determine the aligned and cross-channel matches.
To translate
this joint comparison into retrieved motion, we normalize over anchors:
\begin{equation}
\widetilde{p}_{ia}^{h}=\softmax_{a}(\widetilde{\ell}_{ia}^{h}),
\qquad
\widetilde{\vo}_{i,t}^{h}
=\sum_{a\in\gA}\widetilde{p}_{ia}^{h}\vv_{a,t}^{h}.
\label{eq:trope_readout}
\end{equation}
Here, $\vv_{a,t}^{h}$ is projected from $\vm_{a,t}$. To retain a consistent
structural comparison throughout the sequence, we share queries, keys,
and $r_{ia}$ across frames while updating motion values with the video.
The aggregated features pass to the existing coordinate prediction head. Additional derivations of TRoPE are provided in Appendix~\ref{app:technical}.

\subsection{Sticking-Aware Tail-Risk Training}
\label{sec:training_implementation}

To emphasize persistent local errors that global reconstruction averages
can underweight, we supplement coordinate MSE with vertex-wise CVaR~\citep{cvar}.
We rank vertices by temporal mean error to focus on sustained trajectory
errors. Let
$e_i=T^{-1}\sum_t\|\widehat{\vv}_{t,i}-\vv_{t,i}\|_2$ and let
$\gH_\rho$ contain the $K_\rho=\ceil{\rho N_s}$ largest temporal errors
among the $N_s$ supervised surface locations. The training objective is
\begin{equation}
\Ls
=\Ls_{\mathrm{coord}}
+\reg_{\mathrm{tail}}w(s)
\frac{1}{K_\rho}\sum_{i\in\gH_\rho}e_i,
\qquad
w(s)=\min\!\left(1,\frac{s}{200}\right),
\quad \rho=\reg_{\mathrm{tail}}=0.01.
\label{eq:tail_risk}
\end{equation}
The hardest vertices are selected independently per sample.

Using Motion 3-to-4~\citep{motion324} as our host generator, we fine-tune TRoPE on MeshSticking-4D for 42,600 updates with batch
size eight and a frozen image encoder. We use 12 frames at stride two,
4,096 surface queries, 1,024 FPS anchors, $K=16$ modes per component,
$\gT=\{0.25,1,4,16\}$, and eight fixed logarithmically spaced frequencies
in $[0.5,16]$. AdamW uses a base learning rate of \(3\times10^{-5}\) for the pretrained backbone, while the newly introduced modules use a learning rate of \(9\times10^{-4}\), with cosine decay after 500 warm-up updates~\citep{Adam}.

\section{Experiments}
\label{sec:experiment}

We first evaluate \rev{sticking robustness and motion fidelity},
then examine the positional relation and structural design
behind this behavior.

\subsection{Experimental Setup}
\label{sec:experimental_setup}

\textbf{Datasets.}
We first construct \textbf{MeshSticking-4D}, a benchmark designed to
evaluate sticking artifacts in video-to-4D mesh generation. We combine
Mixamo character meshes with source animations and select clips with
pronounced articulation and close approaches between structurally
distant body parts~\citep{makeitanimatable}.
Each sample contains a reference mesh, a 32-frame mesh sequence,
and monocular videos rendered from 8 synchronized viewpoints.
The benchmark consists of 10080 training samples and 256 test samples.
\added{To evaluate other object geometries, we use Motion324-Subset
from the dataset released with Motion 3-to-4~\citep{motion324},
comprising 1,750 training and 128 test samples across humanoid
characters, animals and other creatures, robots, and rigid or
articulated props. This subset uses 12-frame clips.}
For evaluation on overall generation quality, we additionally use
the public Motion-80 benchmark of Motion 3-to-4~\citep{motion324}
under a reference-mesh-conditioned 16-frame protocol.
Additional details are provided in Appendix~\ref{app:implementation}.

\textbf{Metrics.}
We evaluate motion fidelity using three correspondence-based metrics:
Euclidean Distance (EucD), worst-vertex error (WVE), and framewise maximum
vertex error (FMVE). EucD measures average reconstruction accuracy, while
WVE and FMVE capture vertices with persistent errors and the largest
error in each frame, respectively. To evaluate sticking artifacts, we
identify contact-to-separation events between mesh-distant regions.
Event-endpoint tail error (EETE$_{25}$) measures the hardest 25\% of
event-endpoint vertices, separation failure rate (SFR) measures the
fraction of events that fail to reach the separation threshold, and
mean separation deficit (MSD) measures the shortfall in predicted
separation.
\rev{On Motion-80, we report surface CD, HD95, and F-scores,
together with these six metrics on the same predictions.}
Additional details are provided in Appendix~\ref{app:metrics}.

\textbf{Baselines.}
We compare our approach against five representative methods for
video-guided 4D mesh generation: Motion 3-to-4~\citep{motion324},
Puppeteer~\citep{puppeteer}, ActionMesh~\citep{actionmesh},
R-DMesh~\citep{rdmesh}, and Mesh4D~\citep{mesh4d}.
These methods cover representative feed-forward and topology-preserving
mesh-generation paradigms.
Additional details are provided in Appendix~\ref{app:baselines}.

\added{\textbf{Seam post-processing.}
On Motion324-Subset, we average predicted displacements within each
reference-mesh seam group in each frame to keep duplicate vertices
moving consistently. This training-free step is applied
to TRoPE and Motion 3-to-4 on this subset.}

\begin{figure*}[t]
\centering
\includegraphics[width=\textwidth]{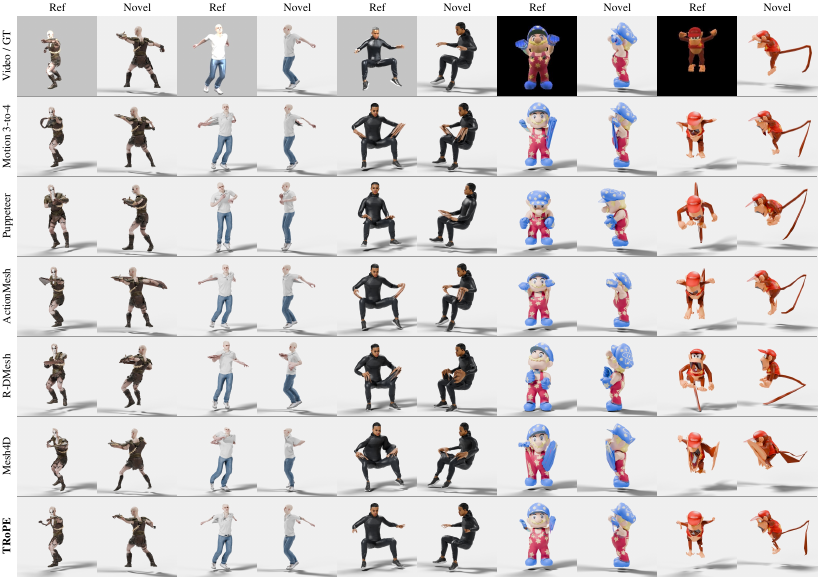}
\caption{
\textbf{Qualitative Comparisons on MeshSticking-4D
\added{and Motion324-Subset}.}
\added{The first three examples are from MeshSticking-4D;
the last two are from Motion324-Subset.}
Each column pair shows reference (Ref) and novel views.
The top row presents the driving video and novel-view ground truth;
the remaining rows compare five baselines and TRoPE.
Compared with the baselines, TRoPE more faithfully recovers articulated
poses while preserving separation between nearby \rev{parts},
reducing sticking artifacts.
}
\label{fig:main_qualitative}
\end{figure*}

\subsection{Main Results}
\label{sec:overall_performance}

\subsubsection{Results on MeshSticking-4D}

We first evaluate the effectiveness of TRoPE through qualitative
comparisons. As shown in Figure~\ref{fig:main_qualitative}
\added{(the first three examples)}, our method
more faithfully reconstructs the driving poses while reducing sticking
artifacts between nearby body parts. Compared with Motion
3-to-4, TRoPE better separates the hands from the
thighs in the third example, avoiding the stretched hands caused by
incorrect coupling between the two regions.
\added{R-DMesh also retains stretching between the hands and thighs
in this example.}

The comparisons also illustrate the importance of recovering motion
together with local geometry. In the second example, R-DMesh and Mesh4D
produce \rev{incorrect arm configurations}, while Puppeteer exhibits
\rev{limited arm extension}. TRoPE more closely matches the
\rev{outstretched-arm pose} and
preserves the separation of the arms from the body. The reference and
novel views provide complementary evidence that our method improves
the reconstructed geometry, reducing the pronounced sticking artifacts.

We quantitatively evaluate TRoPE on the 256 test samples of
MeshSticking-4D. As shown in Table~\ref{tab:main_comparison}, TRoPE
achieves the best EucD, WVE, FMVE, EETE$_{25}$, and SFR among the
compared methods. Relative to Motion 3-to-4, EucD decreases from 59.657
to 54.588, while WVE and FMVE decrease by
39.4\% and 36.2\%, respectively. These reductions show that the
improvement extends beyond average reconstruction accuracy to the
vertices with \rev{persistent or extreme motion errors}.

TRoPE also obtains the best EETE$_{25}$ and SFR. EETE$_{25}$ decreases
from 137.872 to 98.191, SFR from 49.349\% to 34.898\%, and MSD from
35.849 to 29.431.
\added{ActionMesh obtains MSD 27.492 alongside EucD 122.559 and
WVE 305.978, highlighting the importance of evaluating part separation
together with accurate vertex positions.}
Together with the qualitative comparisons, these
results demonstrate that TRoPE improves global motion fidelity while
reducing errors associated with sticking artifacts. More results and
limitations are discussed in Appendix~\ref{app:results}.

\begin{table}[t]
\centering

\caption{
\textbf{Quantitative Comparison on MeshSticking-4D
\added{and Motion324-Subset}.}
We report EucD, WVE, and FMVE for motion fidelity, and EETE$_{25}$,
SFR, and MSD for sticking robustness, with lower values indicating better
performance across all metrics.
\rev{Results are evaluated on 256 and 128 test samples, respectively.}
Bold indicates the best result, and underline indicates the second-best
result.
}
\label{tab:main_comparison}

\begingroup
\fontsize{7.5}{7.3}\selectfont
\setlength{\tabcolsep}{0pt}
\renewcommand{\arraystretch}{1.14}

\begin{tabular*}{\linewidth}{
    @{}l@{\extracolsep{\fill}}*{12}{c}@{}
}
\toprule

\multirow{3}{*}{\textbf{Method}}
& \multicolumn{6}{c}{\added{\textbf{MeshSticking-4D}}}
& \multicolumn{6}{c}{\added{\textbf{Motion324-Subset}}} \\

\cmidrule(lr){2-7}
\cmidrule(lr){8-13}

& \multicolumn{3}{c}{\textbf{Global Motion Fidelity}}
& \multicolumn{3}{c}{\textbf{Sticking Robustness}}
& \multicolumn{3}{c}{\added{\textbf{Global Motion Fidelity}}}
& \multicolumn{3}{c}{\added{\textbf{Sticking Robustness}}} \\

\cmidrule(lr){2-4}
\cmidrule(lr){5-7}
\cmidrule(lr){8-10}
\cmidrule(lr){11-13}

& EucD$\downarrow$
& WVE$\downarrow$
& FMVE$\downarrow$
& EETE$_{25}\downarrow$
& SFR(\%)$\downarrow$
& MSD$\downarrow$
& \added{EucD$\downarrow$}
& \added{WVE$\downarrow$}
& \added{FMVE$\downarrow$}
& \added{EETE$_{25}\downarrow$}
& \added{SFR(\%)$\downarrow$}
& \added{MSD$\downarrow$} \\

\midrule

Motion 3-to-4
& \underline{59.657}
& \underline{285.420}
& \underline{309.116}
& \underline{137.872}
& 49.349
& 35.849
& \added{\textbf{31.129}}
& \added{\underline{136.390}}
& \added{\underline{149.661}}
& \added{\underline{82.078}}
& \added{\underline{70.341}}
& \added{\underline{47.572}} \\

Puppeteer
& 229.205
& 688.001
& 736.039
& 307.215
& 78.49
& 35.330
& \added{108.105}
& \added{261.044}
& \added{283.227}
& \added{159.752}
& \added{100.000}
& \added{66.431} \\

ActionMesh
& 122.559
& 305.978
& 338.738
& 188.150
& \underline{43.73}
& \textbf{27.492}
& \added{85.863}
& \added{211.553}
& \added{236.764}
& \added{127.209}
& \added{\textbf{56.262}}
& \added{\textbf{45.113}} \\

R-DMesh
& 147.973
& 529.012
& 578.764
& 223.051
& 76.80
& 41.657
& \added{95.854}
& \added{239.064}
& \added{267.710}
& \added{148.485}
& \added{80.079}
& \added{54.816} \\

Mesh4D
& 142.397
& 435.920
& 485.914
& 236.312
& 99.44
& 49.565
& \added{79.575}
& \added{212.805}
& \added{241.235}
& \added{138.384}
& \added{99.911}
& \added{66.298} \\

\midrule

\textbf{TRoPE (Ours)}
& \textbf{54.588}
& \textbf{172.874}
& \textbf{197.135}
& \textbf{98.191}
& \textbf{34.898}
& \underline{29.431}
& \added{\underline{33.694}}
& \added{\textbf{120.341}}
& \added{\textbf{134.699}}
& \added{\textbf{79.187}}
& \added{74.433}
& \added{50.375} \\

\bottomrule
\end{tabular*}

\endgroup
\end{table}

\subsubsection{Generalization to Motion-80}
\label{sec:motion80}

We further evaluate TRoPE on Motion-80. As shown in
Table~\ref{tab:motion80}, TRoPE achieves CD 12.238, HD95
34.845, and F-scores of 0.5940 and 0.8656. It outperforms the
reference-conditioned ActionMesh, Mesh4D, Puppeteer, and R-DMesh
adapters \rev{across the four surface reconstruction metrics.}

\added{
We additionally evaluate the six correspondence-based metrics
on the same predictions. TRoPE achieves the best WVE and FMVE among
the compared methods. Relative to Motion 3-to-4, WVE decreases from
99.833 to 91.244, FMVE from 113.487 to 104.805,
and EETE$_{25}$ from 80.509 to 73.505, while CD increases
by 8.7\% and EucD from 24.079 to 26.596.
The average-reconstruction cost is primarily associated with replacing
direct vertex-to-token motion retrieval with two-stage anchor-based
retrieval. This introduces intermediate motion aggregation and a new
vertex--anchor mapping, requiring
the pretrained decoder to adapt to the modified retrieval pathway.
}

\added{
The separation metrics also depend on reference-mesh relation
coverage. On Motion-80, 22.2\% of contact-to-separation events have
at least one endpoint without valid spectral support. For affected
query--anchor pairs, retrieval uses content compatibility without
topological phase modulation, limiting structural guidance for
post-contact separation.
}

\begin{table}[t]
\centering

\caption{
\textbf{Quantitative Comparison on Motion-80.}
\rev{All methods use the same 80 sequences and frames 0--15.
Distance errors are scaled by $10^3$; lower is better except for F-scores.
Bold and underline indicate the best and second-best results.}
}
\label{tab:motion80}

\begingroup
\fontsize{7.5}{7.3}\selectfont
\setlength{\tabcolsep}{0pt}
\renewcommand{\arraystretch}{1.10}

\begin{tabular*}{\linewidth}{
    @{}l@{\extracolsep{\fill}}*{10}{c}@{}
}
\toprule

\textbf{Method}
& CD $\times 10^3\downarrow$
& HD95 $\times 10^3\downarrow$
& F@0.01$\uparrow$
& F@0.02$\uparrow$
& \added{EucD$\downarrow$}
& \added{WVE$\downarrow$}
& \added{FMVE$\downarrow$}
& \added{EETE$_{25}\downarrow$}
& \added{SFR(\%)$\downarrow$}
& \added{MSD$\downarrow$} \\

\midrule

Motion 3-to-4
& \textbf{11.255}
& \textbf{30.926}
& \textbf{0.6386}
& \textbf{0.8891}
& \added{\textbf{24.079}}
& \added{\underline{99.833}}
& \added{\underline{113.487}}
& \added{80.509}
& \added{\underline{50.769}}
& \added{\underline{28.850}} \\

TRoPE
& \underline{12.238}
& \underline{34.845}
& \underline{0.5940}
& \underline{0.8656}
& \added{\underline{26.596}}
& \added{\textbf{91.244}}
& \added{\textbf{104.805}}
& \added{\underline{73.505}}
& \added{58.281}
& \added{33.361} \\

ActionMesh
& 17.957
& 47.193
& 0.5370
& 0.7907
& \added{39.487}
& \added{114.021}
& \added{131.143}
& \added{\textbf{67.429}}
& \added{\textbf{35.294}}
& \added{\textbf{19.263}} \\

Mesh4D
& 34.757
& 111.227
& 0.3524
& 0.5764
& \added{89.296}
& \added{241.831}
& \added{277.744}
& \added{147.798}
& \added{99.729}
& \added{44.581} \\

Puppeteer
& 39.609
& 106.828
& 0.2523
& 0.4500
& \added{90.551}
& \added{219.703}
& \added{248.127}
& \added{146.719}
& \added{86.878}
& \added{37.866} \\

R-DMesh
& 82.277
& 259.940
& 0.1940
& 0.3296
& \added{197.234}
& \added{525.813}
& \added{542.892}
& \added{236.814}
& \added{75.294}
& \added{31.305} \\

\bottomrule
\end{tabular*}

\endgroup
\end{table}

\begin{figure}[t]
\normalsize
\centering
\setlength{\peTableWidth}{0.58\linewidth}
\setlength{\peFigureWidth}{0.40\linewidth}

\sbox{\peTableBodyBox}{%
\begin{minipage}{\peTableWidth}
\centering
\begingroup
\fontsize{7.5}{7.3}\selectfont
\setlength{\tabcolsep}{0.45pt}
\renewcommand{\arraystretch}{1.08}
\begin{tabular*}{\linewidth}{@{}>{\raggedright\arraybackslash}m{0.25\linewidth}@{\extracolsep{\fill}}cccccc@{}}
\toprule
\multirow{2}{*}{\textbf{Method}}
& \multicolumn{3}{c}{\textbf{Global Motion Fidelity}}
& \multicolumn{3}{c}{\textbf{Sticking Robustness}} \\
\cmidrule(lr){2-4}
\cmidrule(lr){5-7}
& EucD 
& WVE 
& FMVE 
& EETE$_{25}$ 
& SFR (\%) 
& MSD  \\
\midrule
\rev{Content-only}
& 55.609 & 208.456 & 230.146 & 118.312 & 44.517 & 32.999 \\
\rev{Abs. Cartesian}
& 55.681 & 208.739 & 231.014 & 116.667 & 45.957 & 33.491 \\
LapPE
& \rev{55.240} & 207.396 & 229.150 & 116.712 & 46.190 & 33.248 \\
\rev{Abs. HKS}
& 55.957 & 207.241 & 229.023 & 117.982 & 44.981 & 33.638 \\
\rev{Euclidean Rel.}
& 55.715 & 207.962 & 231.409 & 118.190 & 46.515 & 33.808 \\
\rev{Geodesic Rel.}
& 55.312 & 191.074 & 215.022 & 102.974 & 37.268 & 30.098 \\
\added{Dist. Bias}
& \added{55.284} & \added{209.695} & \added{231.684}
& \added{118.734} & \added{46.933} & \added{34.333} \\
\added{Kernel Bias}
& \added{\textbf{54.718}} & \added{199.651} & \added{222.856}
& \added{108.819} & \added{42.286} & \added{31.454} \\
\textbf{TRoPE}
& 55.601 & \textbf{181.599} & \textbf{206.476}
& \textbf{101.484} & \textbf{36.152} & \textbf{29.827} \\
\bottomrule
\end{tabular*}
\endgroup
\end{minipage}%
}
\sbox{\peFigureBodyBox}{%
\begin{minipage}{\peFigureWidth}
\centering
\includegraphics[width=\linewidth]{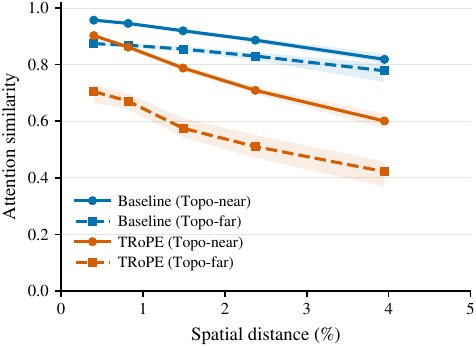}
\end{minipage}%
}
\sbox{\peTableCaptionBox}{%
\begin{minipage}{\peTableWidth}
\normalsize
\setlength{\abovecaptionskip}{0pt}
\setlength{\belowcaptionskip}{0pt}
\captionof{table}{
\textbf{Positional-Encoding Analysis on MeshSticking-4D.}
\added{All variants use the same anchors, PLTA and training protocol
without CVaR. Dist. Bias and Kernel Bias are learned per-head distance
and per-head/per-scale heat-kernel biases.}
Lower is better; bold indicates the best result.
}
\label{tab:pe_analysis}
\end{minipage}%
}
\sbox{\peFigureCaptionBox}{%
\begin{minipage}{\peFigureWidth}
\normalsize
\setlength{\abovecaptionskip}{0pt}
\setlength{\belowcaptionskip}{0pt}
\captionof{figure}{\added{
\textbf{Effect of Topological Relation on Motion Retrieval.}
Attention similarity for spatially matched, topologically near (solid)
and far (dashed) vertex pairs. TRoPE widens the near--far gap relative
to Motion 3-to-4.
}}
\label{fig:retrieval_diagnosis}
\end{minipage}%
}

\setlength{\pePanelHeight}{\dimexpr
  \ht\peTableBodyBox+\dp\peTableBodyBox+
  \ht\peTableCaptionBox+\dp\peTableCaptionBox+5pt\relax}
\setlength{\peOtherHeight}{\dimexpr
  \ht\peFigureBodyBox+\dp\peFigureBodyBox+
  \ht\peFigureCaptionBox+\dp\peFigureCaptionBox+5pt\relax}
\ifdim\peOtherHeight>\pePanelHeight
  \setlength{\pePanelHeight}{\peOtherHeight}
\fi

\noindent
\vtop to \pePanelHeight{%
  \hrule height0pt width\peTableWidth
  \offinterlineskip
  \copy\peTableCaptionBox
  \vskip5pt plus1fill
  \copy\peTableBodyBox
  \vskip0pt
}%
\hfill
\vtop to \pePanelHeight{%
  \hrule height0pt width\peFigureWidth
  \offinterlineskip
  \copy\peFigureBodyBox
  \vskip5pt plus1fill
  \copy\peFigureCaptionBox
  \vskip0pt
}%
\par
\end{figure}

\subsubsection{\added{Results on Other Object Geometries}}

\added{
We further evaluate TRoPE on Motion324-Subset.
As shown in the last two examples of
Figure~\ref{fig:main_qualitative}, Motion 3-to-4 and Mesh4D produce
stretched surfaces between the arms and torso in the fourth example,
which TRoPE substantially reduces. In the last example, Puppeteer and
R-DMesh exhibit pose and shape distortions, while TRoPE better preserves
the articulated geometry.
}

\added{
As shown in Table~\ref{tab:main_comparison}, TRoPE achieves the best
WVE, FMVE, and EETE$_{25}$ on the 128 test samples.
TRoPE reduces WVE and FMVE by 11.8\% and 10.0\% relative to
Motion 3-to-4, and EETE$_{25}$ from 82.078 to 79.187, while EucD changes from 31.129 to 33.694.
These results extend the reduction of persistent and extreme vertex
errors, together with event-endpoint errors, beyond human characters
and MeshSticking-4D.
}

\added{
The relation-coverage constraint discussed in
Section~\ref{sec:motion80} also applies to Motion324-Subset.
Here, seam duplicates account for 30.18\% of test vertices,
and 27.78\% of events have at least one endpoint without valid
spectral support, compared with 0.058\% and 0\%, respectively,
in MeshSticking-4D. This limited coverage explains why the tail-error improvements
do not translate into lower SFR and MSD on this subset.
}

\subsection{Analysis of Topology-Aware Relative Encoding}
\label{sec:pe_analysis}

To understand why TRoPE works, we compare alternative encodings with the
same \rev{Motion 3-to-4} host. CVaR is disabled \added{in Table~\ref{tab:pe_analysis}}
to isolate the query--anchor relation.

\rev{\noindent\textbf{Absolute vs. Relative Encodings.}}
Absolute Cartesian RoPE, \rev{LapPE}~\citep{lappe} and Absolute HKS RoPE
remain close to the Content-only Anchor Attention control.
These results show that describing where a query lies does not
\added{explicitly} establish how it is related to the motion source.

\rev{\noindent\textbf{Euclidean vs. Connectivity-derived Relations.}}
Euclidean Relative RoPE performs similarly to the Content-only Anchor
Attention control, whereas Geodesic Relative RoPE improves quantitative
performance, demonstrating that mesh connectivity cannot be replaced by
spatial proximity and provides crucial structural information for reducing
sticking artifacts.
\rev{TRoPE further reduces WVE from 191.074 to 181.599 and
SFR from 37.268\% to 36.152\%.}
\added{Geodesic distance records shortest-path length, whereas diffusion
features aggregate connectivity across multiple paths and scales.
Even at fixed source vertices and hop distances, the diffusion relation
remains correlated with ground-truth relative motion
(Appendix~\ref{app:diffusion_analysis}), supporting information beyond path length.}

\rev{\noindent\textbf{Additive vs. Rotary Injection.}}
\rev{We compare two additive alternatives: a learnable
per-head distance penalty $-\alpha_h r_{ia}$ with $\alpha_h>0$, and
a learnable per-head/per-scale heat-kernel bias. The latter improves
WVE to 199.651, while rotary modulation reaches 181.599, a further
9.04\% reduction, and lowers SFR by 6.1 percentage points.}
\added{An additive bias supplies a mesh-dependent score offset independent
of the query and key features. Rotary modulation instead lets the relation
alter their feature compatibility, so the same relation can raise or lower
a score according to feature alignment.}
As summarized in Table~\ref{tab:pe_analysis}, the comparison identifies
the key ingredients behind TRoPE: a persistent connectivity-derived
relation and direct modulation of feature compatibility.

\added{\noindent\textbf{Mechanism of motion retrieval.}
We compare attention-distribution similarity between
each query vertex and topologically near (1--2 hops) or far ($>20$ hops)
neighbors at matched spatial distances
(Figure~\ref{fig:retrieval_diagnosis}).
TRoPE increases the mean near--far similarity gap from 0.064 to 0.190;
holding its queries, keys, and anchors fixed, removing rotary modulation
reduces the gap to 0.068. This diagnostic connects the explicit relation
to greater structural discrimination during motion retrieval.}

\subsection{Ablation of Topology-Aware Modeling}
\label{sec:structural_analysis}

We further analyze the structural choices of TRoPE.
\added{Structural comparisons disable CVaR
(Table~\ref{tab:structural_analysis}).}
\rev{Component-only Relation achieves WVE/SFR of 195.604/38.104\%,
while the complete relation reaches 181.599/36.152\%, showing that
within-component connectivity provides information beyond component identity.}
\rev{Multi-scale diffusion improves WVE/FMVE over $\tau=1$, with comparable average and
separation quality. Increasing $K$ improves WVE/FMVE, with smaller gains
beyond $K=16$. Removing PLTA degrades all six metrics, yet the variant
retaining the relation still has lower WVE than Content-only Anchor
Attention with PLTA (194.891 vs. 208.456), demonstrating the complementary
roles of local aggregation and explicit relations.}

\added{Direct + PLTA enriches vertex queries with topology but retrieves
motion without anchors. TRoPE reduces its WVE from 277.633 to 181.599
and SFR from 43.680\% to 36.152\%.
Mesh-supported anchors locate the motion sources, enabling query--source
relations that query-side aggregation alone does not provide.}
\added{Under the same CVaR objective, TRoPE improves all six metrics
over Content-only Anchor Attention, reducing WVE from 204.583 to 172.874
(15.5\%). CVaR further reduces TRoPE's WVE by 4.8\%, compared with
1.9\% for Content-only, supporting complementary gains from topology-aware
retrieval and tail-risk optimization.}

\begin{table}[t]
\centering
\setlength{\tabcolsep}{4pt}
\renewcommand{\arraystretch}{1.10}
\caption{
\textbf{Ablation Analysis on MeshSticking-4D.}
\added{CVaR is off in the upper block and on in the lower block.}
Bold and underline indicate the best and second-best results
\added{within each block}.
}
\label{tab:structural_analysis}
\begingroup
\fontsize{7.5}{7.3}\selectfont
\begin{tabular*}{\linewidth}{@{}l@{\extracolsep{\fill}}cccccc@{}}
\toprule
\multirow{2}{*}{\textbf{Configuration}}
& \multicolumn{3}{c}{\textbf{Global Motion Fidelity}}
& \multicolumn{3}{c}{\textbf{Sticking Robustness}} \\
\cmidrule(lr){2-4}
\cmidrule(lr){5-7}
& EucD $\downarrow$
& WVE $\downarrow$
& FMVE $\downarrow$
& EETE$_{25}$ $\downarrow$
& SFR (\%) $\downarrow$
& MSD $\downarrow$ \\
\midrule
\multicolumn{7}{@{}l}{\added{\textit{Without CVaR}}} \\
Content-only Anchor Attention
& 55.609 & 208.456 & 230.146 & 118.312 & 44.517 & 32.999 \\
\added{Direct + PLTA (no anchors)}
& \added{58.540} & \added{277.633} & \added{300.683}
& \added{122.087} & \added{43.680} & \added{32.238} \\
Component-only Relation
& 56.042 & 195.604 & 218.892 & 110.509 & 38.104 & 30.499 \\
Single-scale Diffusion, $\tau=1$
& \underline{55.594} & 184.127 & 208.673 & 102.306
& \textbf{35.223} & \textbf{29.637} \\
TRoPE, $K=8$
& 55.840 & 186.165 & 210.596 & \textbf{100.347} & 37.128 & 29.828 \\
TRoPE, $K=16$
& 55.601 & \underline{181.599} & \underline{206.476}
& 101.484 & \underline{36.152} & 29.827 \\
TRoPE, $K=32$
& \textbf{55.361} & \textbf{179.965} & \textbf{205.558}
& \underline{100.682} & 36.524 & \underline{29.683} \\
TRoPE w/o Local PLTA
& 56.920 & 194.891 & 219.809 & 107.659 & 37.314 & 31.718 \\
\midrule
\multicolumn{7}{@{}l}{\added{\textit{With CVaR}}} \\
\added{Content-only Anchor Attention + CVaR}
& \added{\underline{55.678}}
& \added{\underline{204.583}}
& \added{\underline{227.164}}
& \added{\underline{117.880}}
& \added{\underline{45.586}}
& \added{\underline{33.713}} \\
\added{\textbf{TRoPE, $K=16$ + CVaR}}
& \added{\textbf{54.588}}
& \added{\textbf{172.874}}
& \added{\textbf{197.135}}
& \added{\textbf{98.191}}
& \added{\textbf{34.898}}
& \added{\textbf{29.431}} \\
\bottomrule
\end{tabular*}
\endgroup
\end{table}

\section{Conclusion}

In this work, we present TRoPE, a topology-aware relative positional
encoding for 4D mesh generation that \rev{reduces sticking artifacts by
making mesh connectivity explicit in query--source motion retrieval}.
TRoPE derives stable multi-scale relations from mesh connectivity and
injects them directly into query--key compatibility through rotary
modulation, while a vertex-wise tail-risk objective emphasizes persistent
high-error regions. We also \rev{introduce} MeshSticking-4D, a benchmark
dedicated to evaluating sticking failures.
\rev{Experiments demonstrate improved motion accuracy and sticking
robustness on MeshSticking-4D, lower tail errors on diverse geometries and transferability to
additional backbone, establishing topology-relative motion retrieval as an
effective approach to reducing sticking in 4D generation.}

\section*{AI Use Statement}

In this work, we used generative AI tools, including large language
models, to assist with research ideation, mathematical derivations
and proofs, code implementation, literature retrieval, and manuscript
preparation. Specifically, these tools supported the discussion and
refinement of research ideas, the development of mathematical
arguments and proof drafts, and the implementation of research code.
We also used these tools to identify relevant prior work, draft and revise portions of the
manuscript, and improve
the clarity and readability of the writing.

The authors reviewed and revised the AI-assisted material incorporated
into this work. In particular, we examined the assumptions and reasoning in
AI-assisted mathematical derivations and proofs, and reviewed and
tested AI-assisted code to check its correctness and consistency
with the methodology described in the paper. The authors made the
final decisions regarding the methodology, implementation,
interpretation of results, and presentation of the work. We take
full responsibility for the final content of this paper and its
associated research artifacts, including all text, mathematical
claims, and code.

\section*{Ethics Statement}

This work investigates topology-aware video-to-4D mesh generation
for animation and geometry reconstruction. MeshSticking-4D is
constructed from Mixamo character and animation assets through
synthetic rendering, with additional evaluation on the existing
Motion-80 benchmark. Generated content should not be presented as authentic
recordings, and downstream applications should respect privacy,
consent, and intellectual-property rights. 
Any use or release of third-party assets should respect their
applicable licenses and redistribution restrictions.
\section*{Reproducibility Statement}

To support reproducibility, Secs.~\ref{sec:preliminaries}--~\ref{sec:topology_relative_rotary} describe the motion
decoding framework, mesh-supported topological relations, and
topology-aware rotary attention, while Sec.~\ref{sec:training_implementation} specifies the
training objective and principal optimization settings.
Appendix~\ref{app:technical} provides additional mathematical
derivations of the multi-scale relation and its rotary formulation.
Appendix~\ref{app:meshsticking} documents the construction of
MeshSticking-4D, including asset collection, motion screening,
rendering, and train--test splits with disjoint character and
source-animation identities. The contact-to-separation event
construction and Motion-80 evaluation protocol are described in
Appendices~\ref{app:contact_events}
and~\ref{app:motion80_protocol}, respectively.
Appendix~\ref{app:plta} details the implementation of PLTA.
Evaluation metrics, baseline implementations, and the configurations
of positional-encoding comparisons and structural ablations are
provided in Appendices~\ref{app:metrics},
\ref{app:baselines}, and~\ref{app:encoding_variants}, respectively.
Additional qualitative comparisons and failure cases are presented
in Appendix~\ref{app:results}. We plan to make our code and benchmark
resources publicly available, subject to the applicable licensing
and redistribution restrictions of third-party assets.

\bibliography{iclr2027_conference}
\bibliographystyle{iclr2027_conference}

\newpage
\appendix
\section{Appendix}

The Appendix provides additional technical details, derivations,
implementation procedures, supplementary experimental settings, and
additional results that complement the main paper. It is organized as follows:
\begin{itemize}
    \item Appendix~\ref{app:technical} presents the derivations and
    theoretical properties of TRoPE.

    \item Appendix~\ref{app:implementation} describes the construction of
    MeshSticking-4D, \added{the Motion324-Subset setting,}
    the evaluation protocols, and additional implementation details.

    \item Appendix~\ref{app:setting} summarizes the supplementary metrics,
    baseline implementations, and the settings of the encoding and ablation experiments.

    \item \added{Appendix~\ref{app:retrieval_analysis} presents
    motion retrieval diagnostics and analyzes TRoPE's structural information.}

    \item \added{Appendix~\ref{app:computational_overhead} reports
    the computational overhead of TRoPE.}

    \item Appendix~\ref{app:results} reports additional qualitative
    results\added{, Mesh4D transfer experiments,} and limitations.
\end{itemize}

\subsection{Derivation and Properties of TRoPE}
\label{app:technical}

In this section, we detail the construction of the topology-relative
relation and its rotary realization introduced in Secs.~\ref{sec:mesh_relative_position} and~\ref{sec:topology_relative_rotary}.
We first expand the multi-scale diffusion relation, then establish its
\rev{invariance to eigenvector sign flips}, and finally
derive how the relation modulates query--key compatibility.

\noindent\textbf{Construction of the Multi-scale Diffusion Relation.}
For each connected component $c$ of the reference mesh, let
$\mA_c$ and $\mD_c$ denote its adjacency and degree matrices. The symmetric
normalized graph Laplacian is

\begin{equation}
\mL_c
=
\mI-\mD_c^{-1/2}\mA_c\mD_c^{-1/2}
=
\mPhi_c\mLambda_c\mPhi_c^{\top}.
\label{eq:app_laplacian}
\end{equation}

We remove the constant mode and retain the non-trivial eigenpairs indexed
by $\gI_c$. Let $\bar{\evlambda}_{c,m}$ denote the eigenvalues
normalized by the median positive retained eigenvalue of component $c$.
For a diffusion scale $\tau$, the normalized spectral response is

\begin{equation}
w_{\tau,c,m}
=
\frac{
\exp(-\tau\bar{\evlambda}_{c,m})
}{
\left\|
\left(
\exp(-\tau\bar{\evlambda}_{c,r})
\right)_{r\in\gI_c}
\right\|_2
}.
\label{eq:app_diffusion_weight}
\end{equation}

The corresponding diffusion feature of a mesh location $i$ is

\begin{equation}
\bm{\psi}_{c,\tau}(i)
=
\left(
\sqrt{w_{\tau,c,m}}\,
\widetilde{\bm{\phi}}_{c,m}(i)
\right)_{m\in\gI_c},
\label{eq:app_diffusion_feature}
\end{equation}

where $\widetilde{\bm{\phi}}_{c,m}$ denotes the retained component-wise spectral
coordinate used by TRoPE~\citep{diffusionnet}. Given a query location $i$ and an anchor location
$a$ on the same component, their normalized relation at scale $\tau$ is

\begin{equation}
r_{ia}^{(\tau)}
=
\frac{
\left\|
\bm{\psi}_{c,\tau}(i)-\bm{\psi}_{c,\tau}(a)
\right\|_2
}{
\max\left\{
\left\|\bm{\psi}_{c,\tau}(i)\right\|_2
+
\left\|\bm{\psi}_{c,\tau}(a)\right\|_2,
\eps
\right\}
}.
\label{eq:app_single_scale_relation}
\end{equation}

Averaging over the fixed set of diffusion scales $\gT$ gives the
final query--anchor relation

\begin{equation}
r_{ia}
=
\frac{1}{|\gT|}
\sum_{\tau\in\gT}
r_{ia}^{(\tau)}.
\label{eq:app_multiscale_relation}
\end{equation}

Thus, $r_{ia}$ compares the structural neighborhoods of the two endpoints.

\noindent\rev{\textbf{Invariance to Eigenvector Sign Flips.}}
\rev{Eigenvectors are defined up to sign~\citep{signnet_basisnet}.
Let $\mQ=\operatorname{diag}(s_m)_{m\in\gI_c}$, with
$s_m\in\{-1,+1\}$, represent independent sign flips.
Centering, column-wise standardization, and symmetric clipping preserve
these sign flips, while the diffusion weights remain unchanged.
The corresponding diffusion features therefore transform as}

\begin{equation}
\bm{\psi}'_{c,\tau}(i)
=
\bm{\psi}_{c,\tau}(i)\mQ.
\label{eq:app_basis_transform}
\end{equation}

For two locations $i$ and $a$, the transformed numerator in
Eq.~\ref{eq:app_single_scale_relation} satisfies

\begin{align}
\left\|
\bm{\psi}'_{c,\tau}(i)-\bm{\psi}'_{c,\tau}(a)
\right\|_2^2
&=
\left\|
\left(
\bm{\psi}_{c,\tau}(i)-\bm{\psi}_{c,\tau}(a)
\right)\mQ
\right\|_2^2 \nonumber\\
&=
\left(
\bm{\psi}_{c,\tau}(i)-\bm{\psi}_{c,\tau}(a)
\right)
\mQ\mQ^{\top}
\left(
\bm{\psi}_{c,\tau}(i)-\bm{\psi}_{c,\tau}(a)
\right)^{\top} \nonumber\\
&=
\left\|
\bm{\psi}_{c,\tau}(i)-\bm{\psi}_{c,\tau}(a)
\right\|_2^2.
\label{eq:app_distance_invariance}
\end{align}

The same argument preserves the two feature norms in the denominator.
Therefore,

\begin{equation}
r_{ia}^{\prime(\tau)}
=
r_{ia}^{(\tau)},
\qquad
r'_{ia}=r_{ia}.
\label{eq:app_relation_invariance}
\end{equation}

\rev{Consequently, the relation used by TRoPE is invariant to
independent eigenvector sign flips.}

\noindent\textbf{Rotary Realization of Relation-conditioned Compatibility.}
Let $\{\evomega_f\}_{f=1}^{F_{\mathrm r}}$ denote the fixed rotary
frequencies. The relation-dependent phase for frequency $f$ is

\begin{equation}
\evtheta_{ia,f}
=
2\pi\evomega_f r_{ia}.
\label{eq:app_rotary_phase}
\end{equation}

For one two-dimensional rotary channel pair, we define

\begin{equation}
\mR(\evtheta)
=
\begin{bmatrix}
\cos\evtheta & \sin\evtheta\\
-\sin\evtheta & \cos\evtheta
\end{bmatrix}.
\label{eq:app_rotary_block}
\end{equation}

Let $\vq=(\evq_1,\evq_2)^{\top}$ and $\vk=(\evk_1,\evk_2)^{\top}$ denote the corresponding
query and key channel pairs. Their relation-conditioned compatibility is

\begin{align}
\vq^{\top}\mR(\evtheta)\vk
&=
(\evq_1\evk_1+\evq_2\evk_2)\cos\evtheta
+
(\evq_1\evk_2-\evq_2\evk_1)\sin\evtheta.
\label{eq:app_rotary_expansion}
\end{align}

The cosine coefficient measures the usual aligned feature similarity,
whereas the sine coefficient captures the cross-channel interaction.
The structural relation determines the two coefficients through the phase,
while the query and key projections determine the feature-dependent
amplitudes. Applying the block rotation to all rotary channel pairs gives

\begin{equation}
\widetilde{\ell}_{ia}^{h}
=
\frac{
\left(\vq_i^{h}\right)^{\top}
\mR(\evtheta_{ia})
\vk_a^{h}
}{
\sqrt{d_h}
},
\label{eq:app_relation_score}
\end{equation}

where $\mR(\evtheta_{ia})$ is block diagonal over the rotary
frequencies. Since each block is orthogonal,

\begin{equation}
\left\|
\mR(\evtheta_{ia})\vk_a^{h}
\right\|_2
=
\left\|\vk_a^{h}\right\|_2.
\label{eq:app_norm_preservation}
\end{equation}

The rotary operation therefore changes the compatibility between the query
and the motion source without rescaling the key. The resulting attention weights and motion readout are

\begin{equation}
\widetilde{p}_{ia}^{h}
=
\frac{
\exp\left(\widetilde{\ell}_{ia}^{h}\right)
}{
\sum_{b\in\gA}
\exp\left(\widetilde{\ell}_{ib}^{h}\right)
},
\qquad
\widetilde{\vo}_{i,t}^{h}
=
\sum_{a\in\gA}
\widetilde{p}_{ia}^{h}\vv_{a,t}^{h}.
\label{eq:app_readout}
\end{equation}

Thus, the topology relation directly affects which motion source is
compatible with a query.

\noindent\textbf{Temporal Reuse and Cross-component Neutrality.}
The relation is computed from the reference connectivity and is therefore
independent of the articulated pose. For every frame $t$,

\begin{equation}
r_{ia}^{(t)}
=
r_{ia}^{\mathrm{ref}}.
\label{eq:app_temporal_reuse}
\end{equation}

This separates the persistent structural relation from the instantaneous motion content.

For locations belonging to different connected components, or for invalid
spectral locations, we define $r_{ia}=0$. It follows from
Eq.~\ref{eq:app_rotary_phase} that

\begin{equation}
r_{ia}=0
\Longrightarrow
\evtheta_{ia,f}=0
\Longrightarrow
\mR(\evtheta_{ia})=\mI,
\label{eq:app_neutral_relation}
\end{equation}

and hence

\begin{equation}
\widetilde{\ell}_{ia}^{h}
=
\frac{
\left(\vq_i^{h}\right)^{\top}\vk_a^{h}
}{
\sqrt{d_h}
}.
\label{eq:app_content_fallback}
\end{equation}

Cross-component pairs therefore retain the ordinary content-based
compatibility. The equations above establish the form used in
Sec.~\ref{sec:topology_relative_rotary}: TRoPE supplies a persistent and \rev{sign-invariant}
query--anchor relation, and injects it into motion compatibility through a
norm-preserving rotary phase.

\subsection{Additional Implementation Details}
\label{app:implementation}

\subsubsection{Details of MeshSticking-4D}
\label{app:meshsticking}

MeshSticking-4D is a benchmark dataset specifically designed to evaluate sticking artifacts. It consists of 10080 training samples and 256 test samples. Each sample contains a reference mesh, a 32-frame mesh sequence, and synchronized monocular views. We construct the benchmark in four steps.

\noindent\textbf{Step 1: Character and Animation Collection.}
Following the Mixamo asset collection and character refinement procedure
provided with Make-It-Animatable\footnote{\url{https://huggingface.co/datasets/jasongzy/Mixamo}}~\citep{makeitanimatable},
we collect 95 rigged, textured characters and 2,453 source animations from
Mixamo. We refine and triangulate the character meshes for motion extraction,
while retaining the textured assets for rendering. We use the 44 characters
that pass the texture and mesh checks to construct the final clip pool.
Animations must contain enough frames to form a 32-frame sequence sampled
at a stride of two.

\noindent\textbf{Step 2: Motion Generation and Candidate Screening.}
We apply each source animation to a character rig and bake its vertex
trajectories. From the resulting motion, we extract 32-frame candidates at
a stride of two, advancing the starting position by 16 sampled frames.
Each candidate is normalized using its sequence bounding box while
preserving its original vertex indexing and face connectivity.

To select motions that expose possible sticking artifacts, we assign
vertices to six coarse body regions: torso, head, left and right arms, and
left and right legs. We sample up to 3,000 vertices and search their
16 nearest spatial neighbors. For pairs from different regions, excluding
the torso--head boundary, we measure their distance over representative
12-frame windows. A pair is counted as a dynamic near-approach when its
minimum distance lies in $(0.06,0.12]$ and its distance changes by at least
$0.08$ within the window. We retain candidates with at least eight such
pairs per window on average and with at least three pairs in no fewer than
half of the evaluated windows. This screening criterion measures the
frequency of challenging close approaches; the contact-to-separation
events used for evaluation are defined separately in the next section.

We additionally require substantial non-rigid motion while maintaining
clean geometry and consistent framing. Specifically, the 95th-percentile
articulation displacement, normalized by body size, must be at least
$0.25$, while normalized centroid span must not exceed $0.50$.
Median and minimum frame occupancy must be at least $0.82$ and $0.60$,
respectively. We limit the 95th-percentile edge strain to $0.20$, the
maximum fraction of collapsed faces to $0.05$, and the normalized
99th-percentile temporal acceleration to $1.00$. Candidates containing
non-finite vertices are excluded.

\noindent\textbf{Step 3: Clip Selection and Rendering.}
Among candidates passing these criteria, we prioritize dynamic
near-approach density and temporal coverage while distributing the selected
clips across characters and source animations. We retain non-overlapping
windows from each character--animation pair. For every selected clip,
we render textured images with Blender Cycles and save the corresponding
mesh trajectory. Images are produced at $256\times256$ resolution. The
camera entries contain eight distinct near-frontal viewpoints. All views use the same mesh trajectory, timestamps,
and sequence-level normalization.

\noindent\textbf{Step 4: Identity-disjoint Split Construction.}
We partition character identities and source-animation identities
independently using a fixed, deterministic ordering. The primary test
split contains 256 clips whose character and source-animation identities
are both absent from the training set. The final training set contains
10080 distinct samples after excluding test identities and incomplete
samples. For training, we use a ground-truth contact-to-separation census
to assign each clip a sampling multiplicity of one, two, or three.
This yields 16223 entries in the training manifest. The resulting
training and test manifests are fixed for all comparisons on
MeshSticking-4D.

\subsubsection{Details of Contact-to-Separation Event}
\label{app:contact_events}

We construct sticking-sensitive events from ground-truth mesh trajectories. For each pair of surface locations, we
compute its reference-mesh connectivity relation and its framewise Euclidean
distance. We retain pairs that are more than 20 mesh hops apart or belong to
different connected components.

A contact-to-separation event is recorded when the pair first enters a
near-contact state with distance below $0.02$ and subsequently reaches a
separated state with distance above $0.10$. The event endpoint is associated
with the corresponding surface locations and their ground-truth trajectories.

The event set is used for every prediction on the testset. This
ensures that EETE$_{25}$, SFR, and MSD measure the behavior of the ground-truth contact-to-separation events.

\subsubsection{Motion-80 Evaluation Protocol}
\label{app:motion80_protocol}

Motion-80 is the public evaluation benchmark released with Motion 3-to-4
\citep{motion324}. We follow its reference-conditioned first-16 protocol.
For each sequence, all methods receive the first-frame reference mesh and
the first 16 driving frames, and predict the corresponding dynamic mesh
sequence.

\rev{For surface metrics, we apply one frame-0 similarity-ICP transform
and reuse it for all evaluated frames.} For each predicted and ground-truth mesh, we sample 10,000 surface
points using the same sampling procedure. We report surface Chamfer Distance
(CD), HD95, and F-scores at normalized distances of $0.01$ and $0.02$.
All methods are evaluated on the same 80 sequences, frames, reference meshes,
and scoring procedure.

\added{We additionally evaluate EucD, WVE, FMVE, EETE$_{25}$, SFR, and MSD
on the same predictions. These six metrics use the original benchmark
coordinates and vertex correspondence without ICP.}

\added{For Motion-80 evaluation, we initialize TRoPE from the released
Motion 3-to-4 checkpoint and fine-tune it with the MSE+CVaR objective
on public Motion324 data. After excluding overlap with Motion-80,
we use 1,646 training sequences and 194 validation sequences.
Fine-tuning runs for 6,750 updates, and Table~\ref{tab:motion80}
reports the checkpoint with the lowest validation coordinate MSE,
selected at update 6,000.}

\subsubsection{Implementation of PLTA}
\label{app:plta}

PLTA is implemented as a local sparse topology-aware attention block
adapted from AnimateAnyMesh++~\citep{animateanymesh++}. Given the reference
mesh connectivity, we construct a sparse graph for the sampled
surface points. For each destination point $i$, the graph contains a
self-edge and up to 16 source points collected from the reference face
graph. Candidate points are ordered by mesh-hop distance, with Euclidean
distance used only to break ties. Missing entries are padded with the
self-edge. For an edge $(i,j)$ with hop distance $h_{ij}$, we add the fixed
topological bias
$b_{ij}=\log(\gamma^{h_{ij}})$, where $\gamma=0.5$.

For input features $\mathbf{x}_i$, PLTA applies pre-normalized multi-head
attention restricted to the local neighborhood
$\mathcal{N}(i)$:
\begin{equation}
\evalpha_{ij}^{(r)}
=
\softmax_{j \in \gN(i)}
\left(
\frac{
\left(
\mW_q^{(r)} \operatorname{LN}(\vx_i)
\right)^{\top}
\left(
\mW_k^{(r)} \operatorname{LN}(\vx_j)
\right)
}{
\sqrt{d_h}
}
+
b_{ij}
\right).
\end{equation}
where $r$ indexes the attention head. The messages from all heads are
concatenated, projected, and added to the input through a residual
connection. A second residual connection then applies a
LayerNorm--linear--GELU--linear feed-forward network. The default
configuration uses 12 attention heads and zero dropout. PLTA is inserted
at the decoder-synchronization stage: it first enriches the dense
reference-point features, after which FPS selects 1024 mesh-supported
anchors from the enriched feature field. The same features are therefore
used to construct dense queries and anchor representations.

\subsubsection{\added{Details of Motion324-Subset}}
\label{app:motion324_subset}

\added{To evaluate other object geometries, we draw Motion324-Subset
from the official Motion 3-to-4 dataset~\citep{motion324}, with 1,750
training and 128 test clips. It covers humanoid characters, animals and
other creatures (e.g., giraffe, monkey, and penguin), robots and mechanical
parts, and rigid or articulated props.}

\added{Motion 3-to-4 and TRoPE start from the same released checkpoint
and are fine-tuned on the same training split for 5,000 updates.
We evaluate the final checkpoints on 12-frame clips using the original
six metrics.}

\added{Both methods use the same training-free seam post-processing.
Seam groups are identified from coincident boundary edges in the reference
mesh. In each frame, predicted displacements are averaged within each
group to keep duplicate vertices moving consistently.}

\subsection{Supplementary Experimental Settings}
\label{app:setting}

In this section, we supplement the evaluation metrics and the implementation details of the baseline, positional encoding and ablation \rev{experiments}. Unless
otherwise specified, all methods are evaluated with the same reference meshes
and driving videos.

\subsubsection{Evaluation Metrics}
\label{app:metrics}

\paragraph{Evaluation on MeshSticking-4D \added{and Motion324-Subset}.}
We evaluate motion fidelity using three correspondence-based metrics.
Let $\widehat{v}_{t,i}$ and $v_{t,i}$ denote the predicted and ground-truth
positions of vertex $i$ at frame $t$, respectively. Euclidean Distance
(EucD) measures the average vertex reconstruction error over all frames and
vertices. Worst-Vertex Error (WVE) is the largest temporal-mean vertex error,
and therefore emphasizes vertices with persistent reconstruction errors.
Framewise Maximum Vertex Error (FMVE) first takes the maximum vertex error
within each frame and then averages these maxima over time.

To evaluate sticking artifacts, we use the contact-to-separation events
defined in Sec.~\ref{app:implementation}. Event-Endpoint Tail Error
(EETE$_{25}$) measures the average \added{temporal-mean vertex} error over the hardest $25\%$ of
event-endpoint vertices. Separation Failure Rate (SFR) is the fraction of
events whose predicted separation does not reach the prescribed separation
threshold. Mean Separation Deficit (MSD) measures the \rev{positive part of the
ground-truth-minus-predicted pair-distance difference, averaged over the
ground-truth contact-to-separation interval and then over events}. Lower values indicate better performance for all six metrics.

\paragraph{Evaluation on Motion-80.}
Following the official Motion-80 protocol, we report surface Chamfer
Distance (CD), the $95$th-percentile Hausdorff Distance (HD95), and the
F-score at normalized distances of $0.01$ and $0.02$. CD and HD95 measure
surface-level reconstruction accuracy, while the F-scores measure the
fraction of points that satisfy the corresponding distance threshold. Lower
CD and HD95 and higher F-scores indicate better performance.
\added{We additionally report the same six correspondence-based metrics
on the same predictions, following the evaluation protocol in
Appendix~\ref{app:motion80_protocol}.}

\subsubsection{Baseline Implementations}
\label{app:baselines}

We provide implementation details for the baseline methods used in our
experiments. Unless otherwise specified, we use the official implementations
or released checkpoints and keep the reference mesh, synchronized driving
view, frame sampling, and output conversion consistent with TRoPE. All
predicted sequences are converted to the common vertex-correspondence format
before evaluation.

\begin{itemize}[
    leftmargin=*,
    labelindent=0pt,
    itemindent=0pt,
    labelsep=0.4em,
    itemsep=2pt,
    topsep=2pt,
    parsep=0pt
]

\item \textbf{Implementation Details of Motion 3-to-4.}
Motion 3-to-4~\citep{motion324} is a feed-forward video-to-4D framework
that encodes a monocular video and a reference mesh into compact motion
tokens and predicts temporally coherent vertex trajectories. We use it as
both the primary host and the content-based control. Specifically, we
initialize the model from the released checkpoint and
fine-tune it on MeshSticking-4D using the same training manifest,
12-frame stride-two windows, and 4,096 surface queries as TRoPE. For the
\rev{main-table baseline}, we disable PLTA, topology-derived features,
topology-relative modulation, and vertex-wise CVaR, while retaining the
\rev{native direct attention from vertex queries to motion tokens}.
\added{The Content-only Anchor Attention ablation instead retains
mesh-supported anchors and local PLTA, removing only the positional relation.} The resulting vertex trajectories
are exported directly in the common reference-mesh correspondence format.

\item \textbf{Implementation Details of Puppeteer.}
Puppeteer~\citep{puppeteer} combines automatic skeleton prediction,
topology-aware skinning-weight estimation, and differentiable
animation optimization. We reproduce the released rigging and skinning
pipelines and reuse the predicted rig for all clips sharing the same
object identity. For each sample, the reference mesh is converted to
Puppeteer's coordinate convention and is animated using the synchronized
12-frame camera-0 video window. Following the released animation
pipeline, we use DPFlow optical flow~\citep{dpflow}, Video-Depth-Anything-Large depth
estimation~\citep{videodepthany}, and CoTracker3 point tracks~\citep{cotracker3}, and optimize the animation for
20 iterations at a rendering resolution of $128\times128$. The final
linear-blend-skinning vertices are transformed back to the benchmark
coordinate system and exported with the original face topology.

\item \textbf{Implementation Details of ActionMesh.}
ActionMesh~\citep{actionmesh} is a temporal 3D diffusion model that first
predicts synchronized time-varying shape latents and then decodes them
into deformations of a reference mesh. \added{For the main-table baseline,}
we use the released mesh-and-video
pipeline with the benchmark reference mesh supplied directly, without
the image-to-3D shape-generation stage, fine-tuning, or test-time
optimization. The input consists of the same camera-0 frame window used
for the other baselines; the 12 selected frames are padded by repeating
the last frame to satisfy the model's 16-frame input interface. We use
the official RMBG preprocessing, 30 temporal denoising steps, guidance
scale 7.5, and bfloat16 inference. The fixed coordinate conversion
between the benchmark and ActionMesh conventions is applied before
inference and inverted before scoring. The decoded mesh sequence is
mapped back to the original vertex ordering and face topology.

\item \textbf{Implementation Details of R-DMesh.}
R-DMesh~\citep{rdmesh} models video-guided animation with a conditional
mesh representation that separates the base mesh, relative trajectories,
and a rectification offset, followed by a rectified-flow generation
stage. For the MeshSticking-4D evaluation, we use the reference-conditioned
R-DMesh pipeline and adapt its mesh decoder on the same weighted training
manifest as Motion 3-to-4. The decoder is initialized from the released
VAE checkpoint and fine-tuned for five epochs with batch size two and
learning rate $10^{-4}$, while the backbone remains frozen. At inference,
we use 4,096 supported mesh points, 512 trajectories, and 64 rectified-flow
steps for the 12-frame benchmark window. Its predicted
vertices are projected to the common reference-mesh correspondence format
before scoring.

\item \textbf{Implementation Details of Mesh4D.}
Mesh4D~\citep{mesh4d} encodes the complete animation into a compact latent
representation and predicts a deformation field conditioned on the video
and the mesh reconstructed from the first frame. We use the released
reference-mesh-conditioned configuration and supply the benchmark
first-frame mesh directly, thereby keeping the output vertex support and
topology fixed across time. The synchronized camera-0 frames are converted
to the RGBA sequence layout expected by the released loader, and the
official Mesh4D checkpoints are used without additional fine-tuning
\added{for the main-table baseline}. The
native image-to-shape generation stage is bypassed because the reference
mesh is fixed by the benchmark protocol. The generated deformation
sequence is exported in the reference-mesh correspondence format and
evaluated with the same metrics as TRoPE.

\end{itemize}

\subsubsection{Encoding and Structural Variants}
\label{app:encoding_variants}

We further provide implementation details for the encoding and
ablation experiments. All experiments use the same Motion 3-to-4 host, checkpoint initialization, data manifests, and training
budget. The common protocol uses 12-frame windows, 4096 query points,
1024 FPS anchors \added{except for Direct + PLTA}, and 42600 optimization steps. CVaR is disabled in these
experiments unless explicitly stated otherwise.

\paragraph{Implementation of positional encoding experiments.}

\begin{itemize}[
    leftmargin=*,
    labelindent=0pt,
    itemindent=0pt,
    labelsep=0.4em,
    itemsep=2pt,
    topsep=2pt,
    parsep=0pt
]

\item \noindent\textbf{Implementation Details of Content-only Anchor
Attention.}
We remove the additional positional relation and retain content-based
compatibility between dense queries and mesh-supported anchors. The anchor
construction and local feature aggregation modules remain unchanged.

\item \noindent\textbf{Implementation Details of Absolute Cartesian RoPE.}
We use the absolute Cartesian coordinates of query and anchor locations as
independent rotary inputs. The rotary frequency budget and attention
architecture are matched to those of TRoPE.

\item \noindent\textbf{Implementation Details of Laplacian Positional
Encoding.}
We inject the Laplacian eigenvectors as absolute decoder
coordinates through additive feature modulation. This experiment corresponds to
LapPE and uses the same spectral resolution as the
default configuration~\citep{lappe}.

\item \noindent\textbf{Implementation Details of Absolute HKS RoPE.}
We construct an absolute scalar heat-kernel-signature field from the reference
mesh and use it as the rotary input for query and anchor features~\citep{hks,hksbase,roformer}. The
rotary channel budget is kept identical to that of TRoPE.

\item \noindent\textbf{Implementation Details of Euclidean Relative RoPE.}
We replace the topology-derived relation with the pairwise Euclidean distance
between each query and anchor. The relation is injected through the same
rotary attention operator as TRoPE.

\item \noindent\textbf{Implementation Details of Geodesic Relative RoPE.}
We compute the shortest-path distance on the reference mesh graph for each
query--anchor pair and use it as the rotary relation. Compared to TRoPE, all remaining components are unchanged.

\item \noindent\added{\textbf{Implementation Details of Learned Per-head
Distance Bias.}
We retain TRoPE's scalar relation $r_{ia}$ and replace rotary modulation
with $\ell_{ia}^{h}=(\vq_i^{h})^{\top}\vk_a^{h}/\sqrt{d_h}
-\alpha_h r_{ia}$, where $\alpha_h=\operatorname{softplus}(u_h)>0$
is learned independently for each head.}

\item \noindent\added{\textbf{Implementation Details of Learned Multi-scale
Kernel Bias.}
We retain the diffusion features and scales, replacing rotary modulation
with $\ell_{ia}^{h}=(\vq_i^{h})^{\top}\vk_a^{h}/\sqrt{d_h}
+\sum_{\tau\in\gT}w_{h,\tau}K_{\tau}(i,a)$.
Here, $K_{\tau}(i,a)=\langle\bm{\psi}_{c,\tau}(i),\bm{\psi}_{c,\tau}(a)\rangle$
for valid endpoints in the same component and zero otherwise;
$w_{h,\tau}$ are learned coefficients for each head and scale.}

\item \noindent\textbf{Implementation Details of TRoPE.}
TRoPE uses the proposed multi-scale mesh-connectivity relative relation as
the rotary modulation of query--key compatibility. This configuration is the
default positional encoding experiment used in the analysis.

\end{itemize}

\paragraph{Implementation of ablation studies.}

\begin{itemize}[
    leftmargin=*,
    labelindent=0pt,
    itemindent=0pt,
    labelsep=0.4em,
    itemsep=2pt,
    topsep=2pt,
    parsep=0pt
]

\item \noindent\textbf{Implementation Details of Component-only Relation.}
We retain connected-component membership as the structural relation while
removing within-component diffusion structure. Cross-component pairs use the
same handling as the other experiments.

\item \noindent\textbf{Implementation Details of Single-scale Diffusion.}
We retain the topology-relative relation but use only the diffusion scale
$\tau=1$, instead of the default multi-scale set.

\item \noindent\textbf{Implementation Details of Spectral Resolution.}
We vary the number of retained spectral modes per connected component over
$K\in\{8,16,32\}$. All other diffusion and rotary settings remain fixed.

\item \noindent\textbf{Implementation Details of TRoPE without Local PLTA.}
We remove the local topology-aware feature aggregation module while retaining
the topology-relative rotary relation, anchor construction, and training
protocol.

\item \noindent\added{\textbf{Implementation Details of Direct + PLTA.}
We retain local PLTA on the vertex queries but remove the intermediate
anchors and vertex--anchor attention. Each enriched vertex query attends
directly to the host motion tokens. CVaR is disabled.}

\item \noindent\added{\textbf{Implementation Details of CVaR Variants.}
We compare Content-only Anchor Attention and TRoPE with and without CVaR,
keeping anchors, local PLTA, and the training budget fixed.
The enabled variants use the objective in Sec.~\ref{sec:training_implementation};
the disabled variants use coordinate MSE alone.}

\end{itemize}

For all encoding and ablation experiments, the reference connectivity is
computed once and reused across frames.

\subsection{\added{Analysis of Topology-Relative Motion Retrieval}}
\label{app:retrieval_analysis}

\subsubsection{\added{Motion Retrieval Diagnosis}}
\label{app:retrieval_diagnosis}

\added{For the diagnostic in Fig.~\ref{fig:retrieval_diagnosis}, we compare
attention-distribution similarity for spatially matched
vertices that are near (1--2 graph hops) or far (more than 20 hops) within
the same connected component. For distributions $p_i$ and $p_j$ over a model's motion
source bank, we define similarity as
$S(i,j)=1-\operatorname{TV}(p_i,p_j)$, where
$\operatorname{TV}(p_i,p_j)=\frac{1}{2}\sum_a|p_{i,a}-p_{j,a}|$.
Similarities are first averaged over heads and frames, then over matched
triplets within each clip and equally across clips. Each model's near--far
gap is computed within its own source bank. Across 4,171 triplets from 235 clips,
the gap is 0.064 for Motion 3-to-4 and 0.190 for TRoPE. With the trained
TRoPE queries, keys, and anchors fixed, removing rotary modulation reduces
the gap to 0.068. This probe isolates the change in retrieval distributions
from rotary modulation while holding the learned features fixed.}

\subsubsection{\added{TRoPE's Advantage over Shortest-Path Distance}}
\label{app:diffusion_analysis}

\added{
Geodesic relative encoding assigns the same relation to vertex pairs
with equal shortest-path lengths, even when their surrounding
connectivity differs. TRoPE captures these differences through
multi-scale diffusion features. To evaluate this additional structural
information, we compare vertex pairs with the same source vertex and
exact graph-hop distance on the largest valid component of each test
character. Ground-truth motion divergence is the temporal mean
difference between the two vertices' displacements from the reference
pose. After centering both the relation and motion divergence within
each source--hop group, their mean Spearman correlation is 0.323
across 24 clips from eight test characters. This shows that TRoPE provides structural distinctions
associated with relative motion that shortest-path distance alone
cannot express.
}

\subsection{\added{Computational Overhead}}
\label{app:computational_overhead}

\added{We benchmark the attention-routing module on one NVIDIA A100 with
$B=1$, $N=4096$ queries, $A=1024$ anchors, 12 heads of dimension 64, 12
frames, and FP32 inputs. Attention weights are computed once for the static
reference relation and reused across frames; value aggregation is performed
once per frame. Table~\ref{tab:attention_cost} reports these routing-module
times. TRoPE and geodesic-relative RoPE have nearly identical cached
12-frame latency (50.22 ms and 50.34 ms), and their per-frame value
aggregation times are 1.04 ms and 1.05 ms. We time the routing operator with
precomputed relations. Reference-mesh preprocessing is performed
once and reused across frames.}

\begin{table}[t]
\centering
\caption{\added{\textbf{Attention-routing latency.}
One-time attention-weight computation, cached value aggregation per frame,
and total latency for 12 frames with uniform caching. Times are in ms.}}
\label{tab:attention_cost}
\setlength{\tabcolsep}{3pt}
\renewcommand{\arraystretch}{1.08}
\begingroup
\fontsize{7.5}{7.3}\selectfont
\begin{tabular*}{\linewidth}{@{}l@{\extracolsep{\fill}}ccc@{}}
\toprule
\added{Method} & \added{Attention weights (once)} &
\added{Value aggregation (per frame)} & \added{12-frame total} \\
\midrule
\added{Content-only} & \added{13.58} & \added{1.35} & \added{29.78} \\
\added{Abs. Cartesian RoPE} & \added{14.04} & \added{1.29} & \added{29.49} \\
\added{LapPE} & \added{13.68} & \added{1.32} & \added{29.57} \\
\added{Abs. HKS RoPE} & \added{14.27} & \added{1.28} & \added{29.65} \\
\added{Euclidean-relative RoPE} & \added{41.05} & \added{1.05} & \added{53.66} \\
\added{Geodesic-relative RoPE} & \added{37.73} & \added{1.05} & \added{50.34} \\
\added{TRoPE} & \added{37.73} & \added{1.04} & \added{50.22} \\
\bottomrule
\end{tabular*}
\endgroup
\end{table}

\subsection{Additional Experimental Results}
\label{app:results}

\subsubsection{More Qualitative Comparisons}
\label{app:more_qualitative}

\begin{figure*}[t]
\centering
\includegraphics[width=\textwidth]{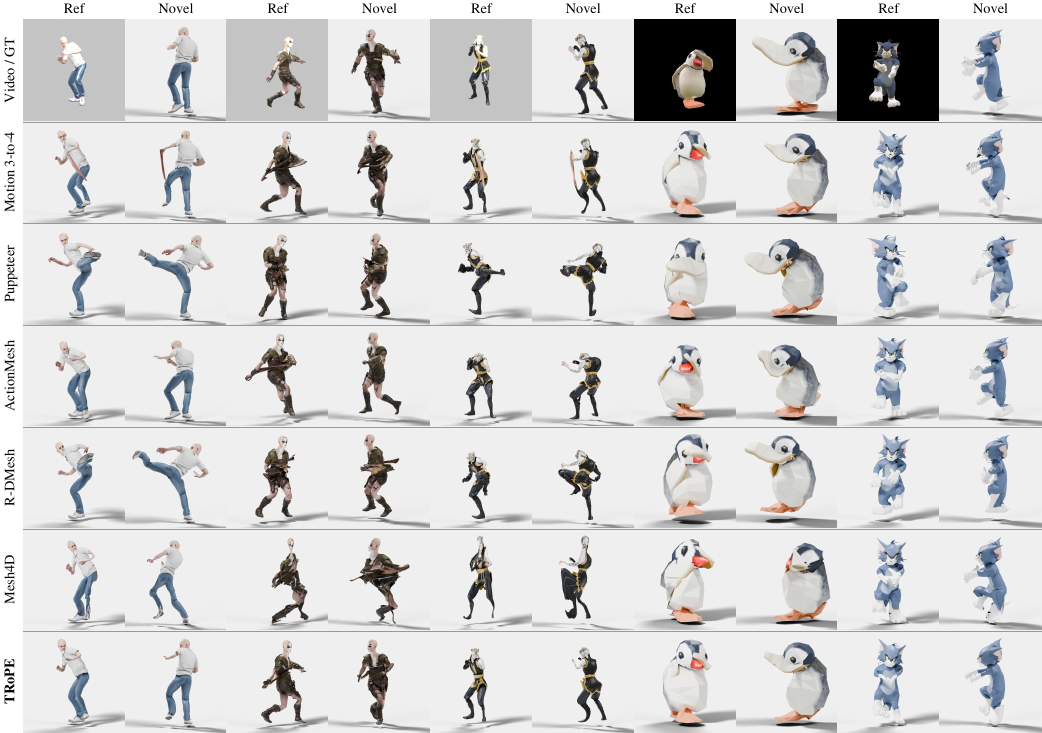}
\caption{\rev{\textbf{Additional qualitative comparisons.}
The first three examples are from MeshSticking-4D and the last two from
Motion324-Subset. Each example pairs reference (Ref) and novel views.
The top row shows the driving video and novel-view ground truth;
the remaining rows compare five baselines and TRoPE.}}
\label{fig:qualitative_1}
\end{figure*}

Beyond the qualitative comparisons presented in the main paper, we provide
\rev{five additional examples in Fig.~\ref{fig:qualitative_1}} to further
evaluate the effectiveness of TRoPE. All methods receive the same reference
mesh and synchronized monocular driving video, and are rendered from
identical reference and novel viewpoints. The results show that TRoPE
\rev{more faithfully} recovers the driving poses while preserving local
surface geometry and part separation.

Across the examples in Fig.~\ref{fig:qualitative_1}, TRoPE maintains coherent
limb configurations and produces consistent silhouettes in both viewpoints.
Several baselines exhibit insufficient pose deformation or distortions in
the arms, legs, and torso, especially under large articulated movements.
\added{In the first example, Motion 3-to-4 stretches the arm toward the
torso, while TRoPE reduces this deformation and better recovers the driving
pose. The second example further shows reduced limb stretching compared
with Motion 3-to-4, R-DMesh, and Mesh4D.}
\rev{The final two examples extend the comparison to Motion324-Subset,
showing a penguin and a stylized cat character. TRoPE better reconstructs their shapes while maintaining good
overall visual quality.}

\subsubsection{\added{Transfer to Mesh4D}}
\label{app:mesh4d_transfer}

\begin{figure}[t]
\centering
\includegraphics[width=\linewidth]{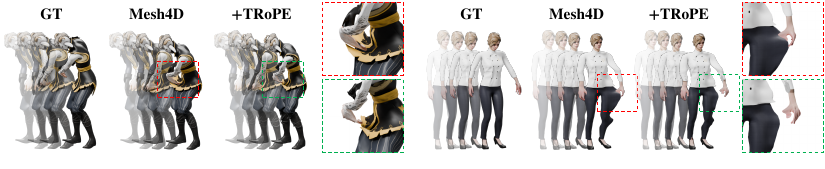}
\caption{\added{\textbf{TRoPE transfer to Mesh4D.}
Two MeshSticking-4D examples compare GT,
the fine-tuned Mesh4D backbone, and the same backbone with TRoPE.
Red and
green insets show the same final-frame region at $2\times$ magnification.}}
\label{fig:mesh4d_transfer}
\end{figure}

\added{We evaluate whether topology-relative motion retrieval transfers to a
different video-to-4D backbone by integrating TRoPE into Mesh4D. We insert a gated topology-relative rotary correction into
Mesh4D's native mesh-query cross-attention, using its 256 canonical
reference vertices to locate motion sources and modulating
cross-component query--source pairs.
Both variants fine-tune the final mesh decoder for 42,600 updates
with coordinate MSE only (CVaR disabled), while all upstream
modules remain frozen. Table
\ref{tab:mesh4d_transfer} reports the paired results under the native
first-16-frame evaluation protocol on 256 test clips. TRoPE yields lower values on all six metrics; EucD remains essentially
unchanged, while WVE and FMVE decrease by 4.0\% and 3.3\%, respectively;
SFR decreases by 22.8 percentage points. The two examples in
Fig.~\ref{fig:mesh4d_transfer} show reduced hand-to-thigh coupling across
distinct examples, supporting transfer beyond the Motion 3-to-4 host.}

\begin{table}[t]
\centering
\caption{\added{\textbf{Transfer to Mesh4D on MeshSticking-4D.}
Both variants use the same training and native first-16-frame
evaluation protocol, with 256 test clips. Lower is
better for all metrics; bold marks the better result within each pair.}}
\label{tab:mesh4d_transfer}
\setlength{\tabcolsep}{3pt}
\renewcommand{\arraystretch}{1.08}
\begingroup
\fontsize{7.5}{7.3}\selectfont
\begin{tabular*}{\linewidth}{@{}l@{\extracolsep{\fill}}*{6}{c}@{}}
\toprule
\added{Method} & \multicolumn{3}{c}{\added{Motion fidelity}} &
\multicolumn{3}{c}{\added{Sticking robustness}} \\
\cmidrule(lr){2-4}\cmidrule(lr){5-7}
& \added{EucD$\downarrow$} & \added{WVE$\downarrow$} &
\added{FMVE$\downarrow$} & \added{EETE$_{25}\downarrow$} &
\added{SFR (\%)$\downarrow$} & \added{MSD$\downarrow$} \\
\midrule
\added{Mesh4D} & \added{130.209} & \added{318.117} & \added{353.858} &
\added{175.858} & \added{99.759} & \added{41.610} \\
\added{Mesh4D + TRoPE} & \added{\textbf{130.204}} &
\added{\textbf{305.285}} & \added{\textbf{342.128}} &
\added{\textbf{172.244}} & \added{\textbf{76.986}} &
\added{\textbf{36.461}} \\
\bottomrule
\end{tabular*}
\endgroup
\end{table}

\subsubsection{\added{Qualitative Comparison of Positional Encodings}}
\label{app:pe_qualitative}

\begin{figure*}[t]
\centering
\includegraphics[width=\textwidth]{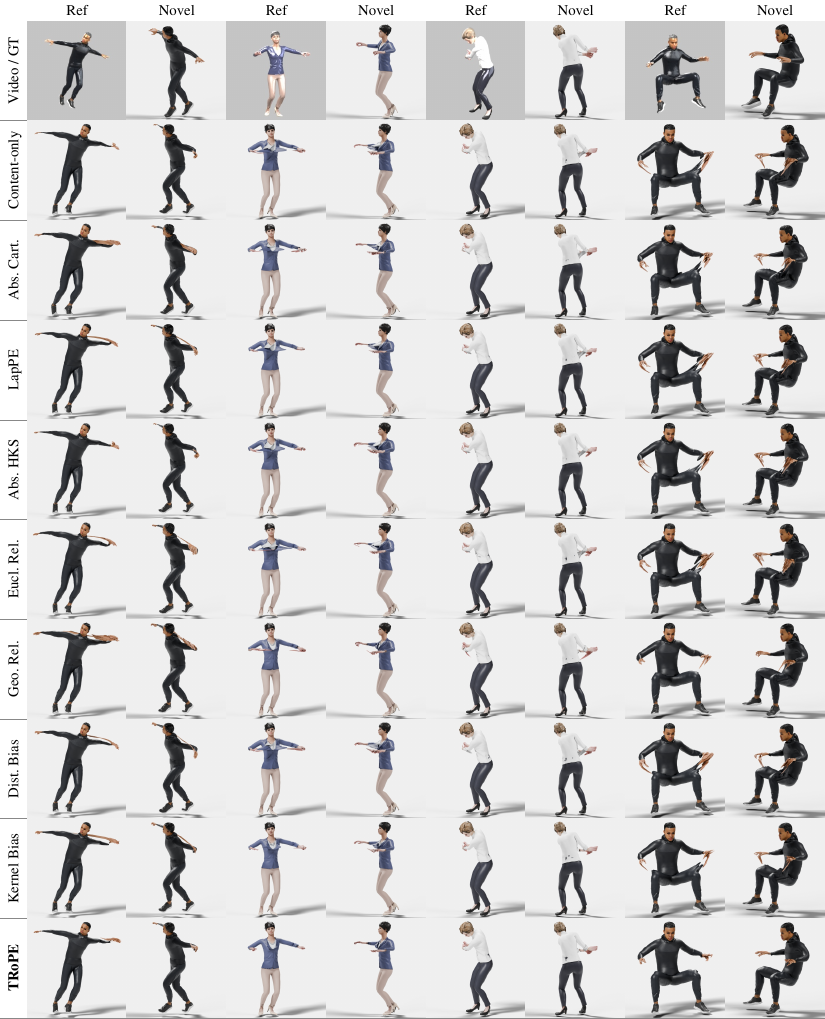}
\caption{\added{\textbf{Qualitative comparison of positional encodings.}
Four MeshSticking-4D examples compare the variants in
Table~\ref{tab:pe_analysis}. Each example pairs reference and novel views of the same frame.
The top row shows the driving video and novel-view ground truth;
the remaining rows compare other positional encodings and TRoPE.}}
\label{fig:pe_qualitative}
\end{figure*}

\added{Figure~\ref{fig:pe_qualitative} compares the positional encodings in
Table~\ref{tab:pe_analysis} on four MeshSticking-4D examples. All variants
retain the same anchors and local PLTA, with CVaR disabled. The comparisons
show how the choice of relation and its injection affect local geometry
during motion retrieval.}

\added{Content-only attention, absolute encodings, and Euclidean-relative
RoPE retain visible stretching between nearby parts. In the second example,
several variants produce a stretched surface between the forearms, while
TRoPE better preserves their separation. Geodesic-relative RoPE improves
part separation but leaves residual hand stretching in the fourth example.
TRoPE reduces this stretching and more closely recovers the hand shape.
These visual differences complement the sticking robustness improvements in
Table~\ref{tab:pe_analysis} and illustrate the benefit of an explicit
diffusion-derived query--source relation.}

\subsubsection{Limitations}
\label{app:limitations}
\label{app:relation_coverage}

\begin{figure*}[t]
\centering
\includegraphics[width=\textwidth]{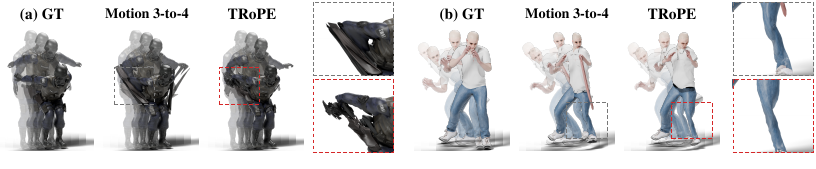}
\caption{\textbf{Limitations.}
(a) In challenging cases, sticking artifacts are mitigated to some extent
but still persist. (b) In some cases, although sticking artifacts are
suppressed, the limbs become deformed.}
\label{fig:limitations}
\end{figure*}

Although TRoPE substantially improves robustness to sticking artifacts, it
still has limitations in challenging motion reconstruction. First, when the
driving video contains severe self-contact, occlusion, or highly ambiguous
motion evidence, the topology-relative relation cannot fully disambiguate
the motion sources, and residual sticking artifacts may persist, as
illustrated in Fig.~\ref{fig:limitations}(a). Second, suppressing
cross-part motion leakage does not guarantee accurate local deformation.
When a large limb displacement exists, the model may preserve part separation while assigning an
inaccurate local trajectory, resulting in limb deformation, as shown in
Fig.~\ref{fig:limitations}(b).

\added{The structural guidance of TRoPE depends on valid relations in the
reference mesh; components with fewer than 16 vertices do not receive spectral features. Table~\ref{tab:relation_coverage} summarizes this coverage
across the three evaluation sets. In Motion-80 and Motion324-Subset,
22.2\% and 27.8\% of events, respectively, have an endpoint without valid
spectral support; duplicate seam vertices account for 52.09\% and 30.18\%
of their test vertices. For query--anchor pairs without a valid relation,
the rotary phase is the identity and motion retrieval falls back to content
compatibility. Thus, fragmented reference connectivity limits where
TRoPE can provide topology-specific guidance.}

\begin{table}[t]
\centering
\caption{\added{\textbf{Reference-mesh relation coverage.}
We report the fraction of vertices in duplicate seam groups, pooled across
test clips, and events with at least one endpoint lacking spectral support.}}
\label{tab:relation_coverage}
\setlength{\tabcolsep}{5pt}
\renewcommand{\arraystretch}{1.08}
\begingroup
\fontsize{7.5}{7.3}\selectfont
\begin{tabular*}{\linewidth}{@{}l@{\extracolsep{\fill}}cc@{}}
\toprule
\added{Dataset} & \added{Duplicate seam vertices (\%)} &
\added{Events with endpoint lacking support} \\
\midrule
\added{MeshSticking-4D} & \added{0.058} & \added{0 / 2,152 (0.0\%)} \\
\added{Motion-80} & \added{52.09} & \added{245 / 1,105 (22.2\%)} \\
\added{Motion324-Subset} & \added{30.18} & \added{1,555 / 5,597 (27.8\%)} \\
\bottomrule
\end{tabular*}
\endgroup
\end{table}

\end{document}